\documentclass{article}
\pdfoutput=1

\usepackage{natbib}
\usepackage{times,latexsym}
\usepackage{url}
\usepackage[T1]{fontenc}
\usepackage[plain]{fancyref}

\usepackage{amsmath}
\usepackage{amsfonts}
\usepackage{subcaption}
\usepackage{mathtools}
\usepackage{booktabs}
\usepackage{tipa}
\usepackage{enumitem}
\usepackage{blkarray}

\newcommand*{\replabel}[1]{\textbf{#1}}
\newcommand*{\featlabel}[1]{\textrm{#1}}

\usepackage{hyperref}
\usepackage{xcolor}
\definecolor{darkblue}{rgb}{0, 0, 0.5}
\hypersetup{colorlinks=true,citecolor=darkblue, linkcolor=darkblue, urlcolor=darkblue}

\begin{document}

\title{Language Embeddings Sometimes Contain Typological Generalizations}

\author{Robert {\"O}stling\thanks{Stockholm University, Department of
Linguistics. Email: robert@ling.su.se} \\ Murathan Kurfal{\i}\thanks{Stockholm University, Department of Psychology, work carried out while at the Department of Linguistics. Email: murathan.kurfali@su.se}}

\maketitle

\begin{abstract}
     To what extent can neural network models learn generalizations about language structure, and how do we find out what they have learned? We explore these questions by training neural models for a range of natural language processing tasks on a massively multilingual dataset of Bible translations in 1295 languages. The learned language representations are then compared to existing typological databases as well as to a novel set of quantitative syntactic and morphological features obtained through annotation projection. We conclude that some generalizations are surprisingly close to traditional features from linguistic typology, but that most of our models, as well as those of previous work, do not appear to have made linguistically meaningful generalizations. Careful attention to details in the evaluation turns out to be essential to avoid false positives. Furthermore, to encourage continued work in this field, we release several resources covering most or all of the languages in our data: (i) multiple sets of language representations, (ii) multilingual word embeddings, (iii) projected and predicted syntactic and morphological features, (iv) software to provide linguistically sound evaluations of language representations.
\end{abstract}

\section{Introduction and related work}

In highly multi-lingual natural language processing (NLP) systems covering hundreds or even thousands of languages, one must deal with a considerable portion of the total diversity present in the world's approximately 7~000 languages. This goes far beyond the standard training resources for most tasks, even after the advent of highly multilingual resources such as the Universal Dependencies Treebanks \citep[][114 languages]{McDonald2013ud} and UniMorph \citep[][110 languages]{SylakGlassman2015unimorph} and unsupervised representation learning models such as multilingual BERT \citep[][104 languages]{Devlin2018bert} and XLM-R \citep[][100 languages]{conneau-etal-2020-unsupervised}.

Thus, for data-driven NLP methods at a scale of more than about a hundred languages one has to rely on different types of data. Although the literature on multilingual NLP is vast, two types of data are generally used for low-resource languages: (a limited selection of) parallel texts, and information on language structure encoded in databases on linguistic typology. Methods based exclusively on parallel texts \citep[prominent early examples being][]{Yarowsky2001inducingpos,Hwa2005bootstrappingparsers} will not be further discussed here, neither will specific ways of \emph{applying} typological information in NLP. For an excellent review of the latter, we refer the reader to \citet{Ponti2019cl}. We are rather interested in the problem of \emph{discovering} typological parameters and obtaining their values for individual languages.

Traditionally, typological databases have been constructed manually\footnote{Although some databases contain automatically computed feature values that can be logically derived from manually specified parameters, AUTOTYP \citep{Bickel2017autotyp} being a prime example of this, we count these as manually constructed since all analysis of language data is performed by humans.}, where collaborations of linguistics researchers classify languages according to a predetermined set of typological parameters. This is a slow\footnote{Case in point: for a number of years we have been looking forward to using the Grambank database (\url{https://glottobank.org/}) in this project, but it has yet to be released.} and costly process, and often leaves large gaps where language documentation is missing or incomplete, or where there has been insufficient researcher time or interest to perform the analysis for some languages. Recent work attempts to exploit correlations between different linguistic features, and between related (or contact) languages to automatically fill some of these gaps \citep{Murawaki2019bayesian,Bjerva2019aprobabilistic}. However, such methods are on their own fundamentally unable to discover cases where a language has diverged from its relatives. While this may be acceptable for NLP engineering purposes, such cases are typically the most interesting linguistically since they provide evidence on the dynamics of language change.

Our main interest is in another approach of obtaining typological parameters, namely estimating typological features using parallel texts. Previous research has demonstrated that a number of typological features can be estimated, using a variety of methods based on parallel texts: semantic categories of motion verbs \citep{Walchli2012lexicaltypology}, word order \citep{Ostling2015wordorder}, colexification patterns \citep{Ostling2016colexification} and tense markers \citep{Dahl2007fromquestionnaires,Asgari2017tense}. All these methods rely on some type of word or morpheme alignment \citep[for an overview, see e.g.][]{Tiedemann2011bitextalignment} combined with manually specified, feature-specific rules. For instance, given a parsed text and word alignments to other languages, one can construct rules for estimating word order properties in those languages. Apart from being a time-consuming process, this also means one has to \emph{know beforehand which features to look for} before hand-crafting rules to estimate their values. In the present work, we focus on the following question:

\begin{center}
    Which typological features can be \emph{discovered} when the system is not told what to look for?
\end{center}

Several authors have attempted to answer this question in recent years, and have reported that everything from phonological to syntactic features can be discovered using a variety of neural models.

The general approach taken is to train a multilingual neural model to perform some NLP task, where the specific language used in a training example is identified using a \emph{language embedding} that is updated during training. This approach was taken by \citet{Ammar2016malopa}, who trained a single dependency parser model using multilingual word representations, with training sentences mixed from seven languages. They found that adding language embeddings improves parsing accuracy, because it allows the model to use the embeddings to adapt to the syntax of each particular language.

While \citet{Ammar2016malopa} showed that language embeddings improve parsing accuracy, they did not investigate whether the language embeddings learned to  \emph{generalize} over languages rather than simply using the embeddings to \emph{identify} each language. A generalization relevant to a parser would be, for instance, whether adjectives tend to precede or to follow the noun that they modify. To say that the model has made this generalization, we would need evidence that the information on adjective/noun order is somehow consistently encoded in the embedding of each language. In contrast, even random language embeddings could serve to identify each language, so that e.g.\ the (random) German embedding is simply used to activate some opaquely coded German syntactic model in the parser's neural network.

\citet{Ostling2017multilm} trained a character-level LSTM language model on translated Bible text from 990 different languages, also conditioned on language embeddings, and showed that the resulting embeddings can be used to reconstruct language genealogies. Similar results have later been obtained using a variety of multilingual neural machine translation (NMT) models that learn either language embeddings \citep{tiedemann2018emerging,Platanios2018contextual}, or language representations derived from encoder activations \citep{Kudugunta2019investigating}, or both \citep{Malaviya2017learning}. In the following we will use the term \emph{language representations} to refer to vector representations of languages, regardless of how they were estimated.

The property that similar languages (i.e.\ languages that share many properties) have similar vector representations is a basic requirement of any useful language representation. It is generally sufficient to improve practical NLP models, especially when evaluated on datasets with many similar languages. However, an aggregate measure of language similarity contains relatively little information. With only similarity information it is impossible to capture the \emph{differences} between otherwise similar languages, for instance when one language differs from its close relatives in some aspect. Even worse, we have no way to learn properties of languages that are not similar to other languages in our data, e.g.\ isolates such as Basque or Burushaski.

Starting with \citet{Malaviya2017learning}, several authors have attempted to probe directly whether the language representations learned by neural models encode the same types of generalizations across languages that have been studied in the field of linguistic typology. \citet{Malaviya2017learning} used logistic regression classifiers to probe whether typological features can be predicted from language representations derived from a multilingual NMT system trained on Bible translations in 1017 languages. They used features from the URIEL database \citep{Littel2017uriel}, which contains typological data sourced from \citet{wals}, \citet{phoible} and \citet{Ethnologue}. Based on their classification experiments, they conclude that their language representations have generalized in several domains of language, from phonological to syntactic features. This finding was later supported by \citet{Oncevay2019towards}, who compared the original representations of \citet{Malaviya2017learning} with a novel set of representations that combined Malaviya's with URIEL features using canonical correlation analysis (CCA).

Similar results have been reported by \citet{Bjerva2018fromphonology}, who use the language embeddings from \citet{Ostling2017multilm} and fine-tune them using specific NLP tasks of several types: grapheme-to-phoneme conversion (representing phonology), word inflection (morphology) and part-of-speech tagging (syntax). Using a k-nearest-neighbors (kNN) classifier for probing, they conclude that typological features from all three domains of language that were investigated (phonology, morphology, syntax) are present in the language representations.

Another, smaller-scale, study on the same topic is that of \citet{He2019syntactic}. They use a denoising autoencoder to reconstruct sentences in 27 languages, using a multilingual dictionary so that the model is presented only with English vocabulary. Based on a syntactic feature classification task, they report that properties of verbal categories, word order, nominal categories, morphology and lexicon were encoded in the language embeddings learned by their autoencoder. They did not see any difference from baseline classification accuracy for features relating to phonology and nominal syntax, a fact that they ascribe to the small amount of languages available for their evaluation.

Finally, we note that a different line of research start from the encoded representations of text, rather than producing explicit language representations. For instance, \citet{chi-etal-2020-finding} use the structural probing technique of \citet{hewitt-manning-2019-structural} to find a syntactic subspace in multilingual BERT encodings of different languages, which allows a direct look at how the model encodes syntactic relations rather similarly across languages. While we find this type of studies interesting, our goal here is different in that we are interested in to what extent neural models given a small set of parameters per language use these to encode typological generalizations.

In summary, the results of previous work
indicate that a range of neural models can learn language representations, which in most cases capture a range of generalizations in multiple domains of language.

A potential problem with the studies listed above is that the method for probing whether a certain feature is captured by some language representations varies, and in several cases is vulnerable to false positives due to the high correlation between features in similar languages. For instance, suppose that a classifier correctly predicts from the Dutch language representation that it tends to use adjective--noun order. Is this because the order of adjective and noun is coded in the language representation space, or because the language representations indicate that Dutch (in the test set) is lexically similar to German (in the training set), which also uses adjective--noun order? Some authors do not control for this correlation between typological features and genealogical or areal connections between languages at all \citep{Oncevay2019towards,He2019syntactic}, others provide the baseline classifiers with genealogical and geographic information \citep{Malaviya2017learning}. \citet{Bjerva2018fromphonology} hold out the largest single language family for each feature as a test set.\footnote{Some details of the evaluation are unclear in the original paper, our summary in this work is also based on personal communication with the authors.} No attempt was made to control for correlations due to language contact.

Given the contradictoy and inconclusive nature of previous work in the area, we here set out to systematically explore our overarching research question of which typological features can be discovered using neural models trained on a massively parallel corpus. The rest of this article is structured as follows.

First, we describe our evaluation framework in \Fref{sec:framework}. In brief, we follow previous work in training classifiers to predict typological features from language representations. To avoid general language similarity from affecting the results, we use a cross-validation scheme that ensures languages in the test fold are not related to, geographically close to, or in a potential contact situation with any of the languages in the training fold. We also provide baselines from lexically derived language representations that are guaranteed \emph{not} to directly code generalizations about language structure.

Second, we describe a diverse set of multilingual neural NLP models that we have implemented (\Fref{sec:languagereps}), based on data derived in various ways from a massively parallel corpus of Bible translations (\Fref{sec:projection}). All models use language embeddings. Since different tasks require analysis at different levels of language, and given the results of \citet{Bjerva2018tracking}, we expect that the language embedding spaces will mainly capture properties relevant to the task at hand.

Finally, we apply our evaluation framework to both our language embeddings, and to several sets of embeddings from previous work (\Fref{sec:experiments}). Surprisingly, we failed to detect any signal of linguistic generalizations in the representations from several previous studies, as well as in most of our own models. We demonstrate multiple ways in which spurious results can be obtained. For some of our models we show that typological features can be predicted with high accuracy, indicating that while neural models \emph{can} discover typological generalizations, they do so less readily than suggested by previous research.

\section{Contributions}

The main contributions of this work are listed below.\footnote{The code required to reproduce the results in this article is available at \url{https://github.com/robertostling/parallel-text-typology} and data is available at Zenodo \citep{ostling_robert_2023_7506220}.}

\begin{enumerate}
    \item A thorough investigation of a number of language representations from previous work as well as newly designed models, including a novel word-level language model that can be trained efficiently on the full vocabularies of thousands of languages.
    \item Publicly available resources derived from parallel texts, for 1295 languages: language representations, multilingual word embeddings, partial inflectional paradigms, and projected token-level typological features relating to word order and affixation type.
    \item A method and publicly available software for detecting typological features encoded into language representations.
\end{enumerate}

\section{Evaluation framework}
\label{sec:framework}

\subsection{Languages and doculects}

In this work, we generally use two levels of granularity, with the following terminology: \emph{languages}, for our purposes identified by a unique ISO 639-3 code, and \emph{doculects}, which is a particular language variety documented in a grammar, dictionary, or text \citep{cysouw-good-2013-doculect}. A typical situation encountered is that for a single language, say Kirghiz (ISO 639-3 code: kir), there are multiple Bible translations and multiple reference grammars. We count this as one language with multiple doculects, which may differ with respect to some features. Here we use the term \emph{doculect} to emphasize that there may be multiple items (e.g.\ Bible translations, word embeddings, language representations) sorting under the same (ISO 639-3) language.

\subsection{Linguistically sound cross-validation}

The basis of our evaluation framework consists of typological feature classification, using constrained leave-one-out cross-validation. Thus, for each feature that we evaluate, the predicted label of a given doculect is obtained from a model that was trained on data from only \emph{independent} doculects. A potential training fold doculect is \emph{non-independent} of the test fold doculect if one or more of the following criteria apply:
\begin{enumerate}
    \item Same family: the training doculect shares top-level family in Glottolog \citep{Hammarstrom2017glottolog31}, including as a special case when they belong to the same language. 
    \item Same macro-area: the test and training doculects belong to the same linguistic macro-area. Although several definitions of macro-areas exist with some differences between them \citep{Hammarstrom2014macroareas}, we rely on the division found in Glottolog \citep{Hammarstrom2017glottolog31}.
    \item Potential long-distance contact: the training and test doculects are listed as potential contact languages in the phoneme inventory borrowings dataset of \citet{segbo2020lrec}.
\end{enumerate}
The first two criteria cover genealogical and areal correlations, respectively. The third criterion  covers some cases that are not directly captured by the previous heuristics, in particular languages such as English and Arabic that are influential globally.

For classification, we follow \citet{Malaviya2017learning} in using L2-regularized logistic regression\footnote{We use a fixed regularization strength $C = 10^{-3}$, and all features (i.e.\ individual language representation dimensions) are scaled to zero mean and unit variance. The use of strong regularization encodes the prior belief that most language representation dimensions are not predictors of a given typological feature. Insufficient regularization in preliminary experiments resulted in strong chance effects from minority class data points.}, as implemented in \citet{scikit-learn}.
The language representations are used directly as features. Our classification models thus contain $k + 1$ parameters, for $k$-dimensional language representations and a bias term.

Naively applying this cross-validation scheme could still lead to problems due to correlations between the representations of related languages, for reasons like lexical similarity. If two large language families (A and B) that share a certain typological parameter P by chance have representations that are similar in some way, and a classifier for a language in family A is trained using (among others) the many languages in B, it will likely predict the parameter P with high accuracy.
The effect of this will be demonstrated empirically in \Fref{sec:naive}.
Since the relationships between languages would be complex to model explicitly, we use family-wise Monte Carlo sampling to estimate classification accuracy and its uncertainty. To compute one sample of the classification accuracy of a given parameter, we uniformly sample one language from each family as the test set. For each language in the test set, we then uniformly sample one language from each family, but only among the \emph{independent} languages (as defined above) to form the corresponding training fold. This procedure is repeated 401 times, yielding 401 samples of the classification accuracy for the given parameter and language representations.
For each classification accuracy sample $a_c$, we also collect a paired baseline accuracy sample $a_b$ by randomly shuffling the training labels. This allows us to verify that the baseline behaves like a binomial distribution with $p = 0.5$.

\section{External resources}
\label{sec:data}

In this section we describe the data we use from external sources, leaving data sets produced by us as part of this work to \Fref{sec:projection}, and the typological databases used for evaluation to \Fref{sec:evaluationdata}.

As the main multilingual resource, we use a corpus of Bible translations crawled from online sources \citep{Mayer2014bible}. In the version used by us, it contains 1846 translations in 1401 languages. This discrepancy is due to some languages having multiple translations. Here we define \emph{language} as corresponding to a unique ISO 639-3 code, while \emph{doculect} refers to the language documented in a single translation. We exclude partial translations with fewer than 80\% of the New Testament verses translated. The count of verses varies somewhat between different traditions, but we compute a canonical set of verses, defined as all verses that occur in at least 80\% of all translations, in total 7912 verses. We also exclude a few translations without suitable word segmentation. A total of 1707 translations in 1299 languages satisfy these criteria. 
For languages that we intend to use as source languages for annotation projection, we manually choose a single \emph{preferred translation} per language. We apply the following criteria for the doculect in the preferred translation, in order of priority:
\begin{itemize}
    \item The doculect should be as close as possible to the modern written variety of the language, in order to match the external resources. This typically means excluding old translations, based on metadata on publication year in the corpus.
    \item The translation should be as literal as possible, without extensive added elaborations or divergences.
    \item The translation should cover the Old Testament part of the Bible, in order to maximize the amount of parallel data to those target texts that contain both the New Testament and the Old Testament.
\end{itemize}
A total of 43 such translations are chosen. These are only used as sources for annotation projection, which brings the number of available \emph{target} translations down to 1664, in 1295 languages. Note that most (39) of the 43 languages used as sources have multiple translations, which means that the non-preferred translations are used as targets during annotation projection (discussed further in \Fref{sec:projection}).

For the word embedding projection (\Fref{sec:embedding-projection}) we use the multilingual word embeddings of \citet{Smith2017offlinebilingual}, trained on monolingual Wikipedia data and aligned into a multilingual space using the English embeddings as a pivot. We chose the 32 languages with the highest word translation accuracy in the evaluation of \citet{Smith2017offlinebilingual}, and refer to these embeddings as \emph{high-resource embeddings} below.

For dependency relation and part of speech tag projection, we lemmatize, PoS-tag and parse Bible translations using the multilingual Turku NLP Pipeline \citep{kanerva2018turkuparser}. A total of 35 languages in the Bible corpus are supported by this model, and the preferred translation in each of these languages is annotated with lemmas, PoS tags and dependency structure following the Universal Dependencies framework \citep{McDonald2013ud}.

For concept labels (\Fref{sec:ids}) we rely on the Intercontinental Dictionary Series, IDS \citep{ids} and its connection to the Concepticon list of semantic concepts \citep{concepticon}. This is a collection of digital lexicons for 329 languages or varieties, of which 25 are supported by the TurkuNLP lemmatizer. Since the IDS contains only citation forms, we only use the lexicons for these 25 languages.

\section{Multi-source projection of information}
\label{sec:projection}

We now turn to several types of resources that we have produced, for use as training data and evaluation. These resources rely on aligning words between a large amount of pairwise translations in the Bible corpus, and \Fref{sec:alignment} below describes an efficient method for performing this task.

\subsection{Subword-based word alignment}
\label{sec:alignment}

Word alignment is performed using subword-level co-occurrence statistics.\footnote{We initially experimented with using a more complex two-step procedure, where subword-level co-occurrence alignment was used to compute Dirichlet priors for a Gibbs sampling aligner based on a Bayesian IBM model \citep{Ostling2016efmaral}. In spite of significantly larger computational cost we did not observe any substantial differences when evaluated on the task of inferring word order properties, as in \fref{sec:wordorderstatistics}.} Since the typical translation pair is unrelated and the languages have very different morphological properties, we prefer this method over word-based alignments. The alignment score of two items, $w$ (from language $L_1$) and $u$ (from language $L_2$), compares two models for explaining co-occurences between $w$ and $u$:
\begin{itemize}
    \item $M_1$: Whether $w$ and $u$ occur in a given Bible verse is decided by draws from two independent Bernoulli distributions.
    \item $M_2$: Whether $w$ and $u$ occur in a given Bible verse is decided by a draw from a categorical distribution with four outcomes ($w$ only, $u$ only, both, or neither).
\end{itemize}
In order to estimate our belief in $M_2$, a systematic co-occurrence,\footnote{Note that $M_2$ simply describes that $w$ and $u$ are not independently distributed, which could also mean that they have a complementary distribution. Since we still align on a token level, requiring instances of $w$ and $u$ to be present in the same verse, this is not a problem in practice.} we multiply our prior belief in $M_2$ with the Bayes factor of $M_2$ over $M_1$. Since a mopheme in one language is translation equivalent to (very) approximately one morpheme in the other language, we use a prior of $1/V$ where $V$ is the total number of unique subwords in $L_1$. We define a \emph{subword} as any substring $w$ of a token which has a higher frequency than any substring $w'$ containing $w$. For instance, if the substring `Jerusale' has the same frequency as `Jerusalem', only the latter will be added to the subword vocabulary.

We use uniform Beta and Dirichlet priors, respectively, for $M_1$ and $M_2$. The resulting alignment score can thus be computed as follows, by combining the prior with the log-Bayes factor $\mathrm{BF}(M_2/M_1)$:
\begin{equation}
\label{eq:alignment}
\begin{aligned}
    s(w,u) &= \log \frac{1}{V} + \mathrm{BF}(M_2/M_1) \\
    &=\log \frac{1}{V} \\
    &+ \log P(<n_w-n_{wu}, n_u-n_{wu}, n_{wu}, n-(n_w+n_u-n_{wu})> | \mathbf{1}) \\
    &- \log P(<n_w, n-n_w> | \mathbf{1}) 
\end{aligned}
\end{equation}
where $n$ is the total number of verses that occur in both the $L_1$ and $L_2$ translations, $n_w$, $n_u$ and $n_{wu}$ the number of verses containing $w$, $u$ and both, respectively.
The Dirichlet-multinomial (and its special case, the Beta-binomial) likelihood function is given by
\begin{equation} P(\mathbf{x} | \mathbf{\alpha}) =
\frac{\Gamma(\sum_i \alpha_i)}{\Gamma(\sum_i (x_i + \alpha_i))} \cdot \prod_k \frac{\Gamma(x_k + \alpha_k)}{\Gamma(\alpha_k)}
\end{equation}
Note that \Fref{eq:alignment} gives a type-level score. In order to get token-level alignments, we greedily align each token in $L_1$ to the highest-scoring token in the corresponding verse of $L_2$. The score $s(w,u)$ is then used as a threshold to filter out tokens that should be left unaligned. In our experiments, we use the criterion $s(w,u) \geq 0$, in other words that $H_2$ should be at least as credible as $H_1$. In addition, we use a few empirically determined thresholds for additional filtering: the log-Bayes factor $\mathrm{BF}(H_2/H_1)$ must be greater than $0.2 n_{wu}$ \emph{and} greater than $\min(100, 0.7 n_{wu})$.

\subsection{Multilingual word embeddings}
\label{sec:embedding-projection}

The multilingual high-resource embeddings described in \Fref{sec:data} cover only 32 languages in our sample, which corresponds to less than 3\% of the languages in the Bible corpus. In order to obtain multilingual word embeddings for all languages we study, we perform word alignment as described above, followed by naive multi-source projection by averaging over the embeddings of all aligned tokens. We use only one translation per language as source. When multiple translations exist for a given language, we have aimed to choose the one closest matching the relatively modern language that the high-resource embeddings have been trained on. In total, we project embeddings to 1664  translations in 1295 different languages.

\subsection{Semantic concepts}
\label{sec:ids}

In order to obtain annotations of semantic concepts for each language, we use lexicons in 25 languages from the Intercontinental Dictionary Series (IDS, \citealt{ids}) which was described further in \Fref{sec:data}. In total 329 languages are available in the IDS, but we only use a subset of 25 languages where we have access to accurate lemmatizers. Each IDS lexicon entry is connected to a common inventory of semantic concepts from the Concepticon \citep{concepticon}, such as \textsc{tree}, \textsc{water} and \textsc{woman}. For each token we assign any concepts that are paired with its lemma in the IDS database. We choose a single semantic concept of a target-text token using a simple majority vote among all the aligned source text tokens, as long as at least 20\% of source texts agree on the given concept label. This procedure is identical to the PoS and dependency relation projection described in \Fref{sec:wordorderstatistics}.

\subsection{Paradigms}
\label{sec:paradigms}

For our reinflection model (\Fref{sec:reinflection}) as well as the affixation type evaluation data (\Fref{sec:affixation}) we need examples of (partial) inflectional paradigms for each language. We approximate these using a combination of the PoS projections (\Fref{sec:wordorderstatistics}) and semantic concept projections (\Fref{sec:ids}). To obtain paradigm candidates for a given language, we perform the following heuristic procedure:
\begin{enumerate}
    \item For each semantic concept, find the PoS tag most commonly associated with it.
    \item Among the word forms with the given projected concept label and PoS tag, perform hierarchical clustering using mean pairwise normalized Levenshtein distance as the distance function.
    \item Select clusters with at least two members, with at least one word form above 4 characters in length, and with a mean pairwise normalized Levenshtein distance below $0.3$.
\end{enumerate}
The normalized Levenshtein distance used is $d(s_1,s_2)/(|s_1|+|s_2|)$, where $d$ is unweighted Levenshtein distance \citep{Levenshtein1966}.
This method also means that we have an estimate of the part of speech for each paradigm, and in the present work we use this information to restrict our study to only noun and verb paradigms. Any part of speech with less than 50 partial paradigms identified is considered to lack inflection. Such low counts have been empirically determined to arise from noise in the alignment procedure.

\subsection{Affixation type}
\label{sec:affixation}

Using noun and verb paradigms estimated in \Fref{sec:paradigms}, we can guess the proportion of prefixing and suffixing inflections by the following procedure. First, we sample 1000 word pairs for each part of speech from each Bible translation, such that the word in each pair comes from the same paradigm, e.g.\ \textit{annotate--annotating}. We then use the Levenshtein algorithm to compute the positions of the edit operations between the two words. If all operations are performed on the first half of each word, we count the pair as prefixing. If all operations are performed  on the second half, we count it as suffixing. Otherwise, we count it as neither.

We evaluate the result of this heuristic by comparing against \citet{wals-26}. To investigate the effect of avoiding ambiguous cases, we consider two cases. In the \textbf{Non-exclusive} condition, prefixing languages are those classified as weakly or strongly prefixing, or as being equally prefixing and suffixing. In the \textbf{Exclusive} condition, only languages which are weakly or strongly prefixing are counted, and all other languages (with either little affixation, or equally prefixing and suffixing) are discarded from the analysis. We define suffixing similarly.

\Fref{tab:projected-word-order} shows the level of agreement with \citet{wals-26}. The table presents accuracy as well as F$_1$ scores. The F$_1$ score presented is the mean of both classes, positive and negative. Since our heuristic classifies all languages as either prefixing or suffixing, we mainly consider the \textbf{Exclusive} condition. Since our sample is strongly biased towards a few large language families, we focus on the \textbf{Family}-balanced scores which weighs each doculect so that all top-level language families receive unit weight. For \textbf{Language} weighting, each ISO 639-3 language code receives unit weight, which is more easily comparable to previous work.  We here achieve a family-balanced accuracy of 85.6\% and a mean F$_1$-score of 0.798. This result is pulled down mainly by the low performance for identifying prefixing languages (recall 74\% and precision 65\%).

Concurrent work has confirmed that automatic estimation of affixation type is quite challenging \citep{hammarstrom-2021-measuring}, for a variety of reasons including the difficulty of identifying productive patterns, and differentiating between inflectional and derivational morphology. \citet{wals-26} specifically concerns inflectional morphology, whereas our method is not able to fully separate inflectional morphology from derivational morphology, or affixes from clitics. We also note that while \citet{wals-26} is counting the number of categories marked by affixes, we are counting the number of word forms with a given affix. Given the high agreement reported above, we do however consider our approximation to be good enough for further investigation.

\subsection{Word order statistics}
\label{sec:wordorderstatistics}

The typological databases used in our evaluation (described further in \Fref{sec:evaluationdata}) have two shortcomings: they are sparse and categorical. Through multi-source projection is it possible to obtain reliable word order statistics \citep{Ostling2015wordorder} for all of the languages in our data, which makes us able to compare how well our data (Bible translations) matches the typological databases used. It is also possible to use the projected features as classifier training data in the evaluation, and as a reference point for analyzing the classification results.

We use the token-level word alignments between each of the 35 Universal Dependencies-annotated translations (see \Fref{sec:data}) and the 1664 low-resource translations to perform multi-source projection of PoS tags and dependency relations. Note that for our purposes we do not need to produce full dependency trees, so dependency links are projected individually.\footnote{We have experimented with using maximum spanning tree decoding to ensure consistency, but did not observe any improvement in word order estimation.} Each PoS tag, dependency head or dependency label needs to be projected from at least 20\% of the available source texts. Otherwise the projection is discarded, as a means of filtering out inconsistent translations and poorly aligned words.

For each language we count the proportion of head-initial orderings for each dependency label and head/dependent PoS combination, to obtain a word order feature matrix covering all languages. The projected word order properties are listed in \Fref{tab:projected-word-order}. For instance, the well-studied typological parameter of object/verb order (where the object is headed by a noun) is captured by the head-initial ratio of \textsc{noun}/\textsc{propn} $\xleftarrow{\text{obj}}$ \textsc{verb} relations. A value of 0 would indicate strict object--verb order, while 1 indicates strict verb--object order, and 0.5 indicates that both ordering are equally frequent on a token basis.

A fundamental assumption in annotation projection is that grammatical relations are the same across translation equivalent words in different languages. While this does not hold in general, several things can be done to make the approximation closer. One source of disagreement is the differences in part-of-speech categories across languages. By focusing on core concepts of each category we can decrease the number of cases where translation equivalents participate in different syntactic relations because they belong to different parts of speech.
\citet[pp 3--7]{Dixon1982adjectives} showed that a small set of concepts are most likely to be lexicalized across languages as true adjectives, that can be used attributively. When estimating adjective/noun order, we limit ourselves to this set.\footnote{We used the following Concepticon labels to define core adjectives: \textsc{strong}, \textsc{high}, \textsc{good}, \textsc{bad}, \textsc{small}, \textsc{big}, \textsc{new}, \textsc{young}, \textsc{old}, \textsc{beautiful}.} \citet{Ostling2019alt} showed that restricting the category of adjectives when projecting relations across Bible texts results in a much closer match to the adjective/noun order data from \citet{wals-87}, as compared to using the Universal Dependencies \textsc{adj} tag. A similar approach was taken for numerals, where only the numerals 2--9 were chosen. This range was chosen to ensure that for the vast majority of languages with a numeral base of 10 or above \citep{wals-131}, only atomic numerals would be chosen and the problem of complex numeral constructions for higher numbers can be avoided \citep{Kann2019}. The word for the numeral 1 is often used for other functions (cf the article \emph{ein} in German), which would have posed additional challenges for accurate parsing and annotation projection.

One problem with using the core adjective concepts of Dixon is that these sometimes stand out from the larger class of adjectives with respect to  word order. A familiar example is the Romance languages, where many of the core adjectives use adjective--noun order instead of the more productive noun--adjective order, but examples are spread across the world \citep{Ostling2019alt}. An alternative method would have been to automatically separate attributive constructions from other types of constructions, but this is a complex problem beyond the scope of this work.

For nouns and verbs, we simply use the Universal Dependencies \textsc{noun} and \textsc{verb} tags, respectively. The high level of agreement with verb/object order data from \citet{wals-81} indicates that this approximation is accurate.

\Fref{tab:projected-word-order} shows the featues we project, their definition in terms of projected Universal Dependencies relations, and the level of agreement with WALS and Ethnologue (as aggregated and binarized by the URIEL database). All values are binarized so that a majority of head-initial projected relations give the value 1, otherwise 0. When multiple URIEL features describe the same phenomenon, the one with the head-initial interpretation is chosen (e.g.\ S\_OBJECT\_AFTER\_VERB) for consistency. Projections are summarized at the level of a doculect, in our case a single Bible translation. Each language may have multiple translations, and a given language family may be represented by multiple languages. As mentioned in \Fref{sec:affixation} above, we report results by weighting so that either languages or language families are given identical weight. We consider the latter to be more informative, since it approximates the expected performance on a newly discovered language from a previously unknown family. Language-weighted numbers are included for ease of comparison with previous work, and to show the effect of using a language sample biased towards some families. We consider mean F$_1$ scores to be more informative, since several of the features are heavily biased towards one class which often leads to inflated accuracy figures (e.g.\ subject/verb order) for methods biased towards the majority class.

Overall, there is a high level of agreement  between the projected features and the classifications from WALS and Ethnologue. Looking at the \textbf{Exclusive} condition, which we use in our later experiments, the family-wise mean F$_1$ scores are 0.8 or above for all features except subject/verb and oblique/verb order. A thorough error analysis is beyond the scope of this work, but some previous work exists on projected typological parameters. \citet{Ostling2019alt} investigated projected adjective/noun order and found a varied number of causes for disagreements with typological databases, including coding errors in the databases themselves, and differences between the Bible translation and reference grammar doculects. In \Fref{sec:error-analysis}, we investigate a number of cases where the projections and databases disagree and find that those can be explained by languages with mostly free word order having been manually classified as having some dominant word order. This is also in line with the findings of \citet{choi-etal-2021-investigating}, who compared quantitative word order data from Universal Dependencies treebanks with WALS classifications.

\begin{table}[tb]
    \centering
    \footnotesize
    \begin{tabular}{llrrrr}
        & & \multicolumn{2}{c}{\textbf{Language}} & \multicolumn{2}{c}{\textbf{Family}} \\
        \textbf{Label} & \textbf{Definition} & \textbf{Accuracy} & \textbf{F$_1$} & \textbf{Accuracy} & \textbf{F$_1$} \\
        \toprule

\midrule
\multicolumn{6}{c}{\textbf{Non-exclusive}} \\
\midrule
Object/verb order & \textsc{noun}/\textsc{propn} $\xleftarrow{\text{obj}}$ \textsc{verb} & 94.7\% & 0.945 & 87.0\% & 0.866 \\
Oblique/verb order & \textsc{noun}/\textsc{propn} $\xleftarrow{\text{obl}}$ \textsc{verb} & 76.1\% & 0.640 & 71.7\% & 0.637 \\
Subject/verb order & \textsc{noun}/\textsc{propn} $\xleftarrow{\text{nsubj}}$ \textsc{verb} & 81.4\% & 0.689 & 84.3\% & 0.578 \\
Adjective/noun order & \textsc{adj*} $\xleftarrow{\text{amod}}$ \textsc{noun} & 81.7\% & 0.799 & 78.7\% & 0.778 \\
Relative/noun order & \textsc{verb} $\xleftarrow{\text{acl}}$ \textsc{noun} & 91.9\% & 0.850 & 86.5\% & 0.797 \\
Numeral/noun order & \textsc{num*} $\xleftarrow{\text{nummod}}$ \textsc{noun} & 92.6\% & 0.926 & 89.2\% & 0.889 \\
Adposition/noun order & \textsc{adp} $\xleftarrow{\text{case}}$ \textsc{noun} & 94.8\% & 0.947 & 95.8\% & 0.955 \\
Prefixing & Prefixes $\geq$ 50\% & 80.9\% & 0.766 & 83.5\% & 0.804 \\
Suffixing & Suffixes $\geq$ 50\% & 70.7\% & 0.646 & 71.2\% & 0.619 \\
\midrule
\multicolumn{6}{c}{\textbf{Exclusive}} \\
\midrule
Object/verb order & \textsc{noun}/\textsc{propn} $\xleftarrow{\text{obj}}$ \textsc{verb} & 95.8\% & 0.957 & 88.6\% & 0.880 \\
Oblique/verb order & \textsc{noun}/\textsc{propn} $\xleftarrow{\text{obl}}$ \textsc{verb} & 76.1\% & 0.640 & 71.7\% & 0.637 \\
Subject/verb order & \textsc{noun}/\textsc{propn} $\xleftarrow{\text{nsubj}}$ \textsc{verb} & 86.8\% & 0.735 & 92.3\% & 0.673 \\
Adjective/noun order & \textsc{adj*} $\xleftarrow{\text{amod}}$ \textsc{noun} & 85.8\% & 0.846 & 85.5\% & 0.850 \\
Relative/noun order & \textsc{verb} $\xleftarrow{\text{acl}}$ \textsc{noun} & 92.4\% & 0.861 & 90.4\% & 0.851 \\
Numeral/noun order & \textsc{num*} $\xleftarrow{\text{nummod}}$ \textsc{noun} & 95.1\% & 0.951 & 92.0\% & 0.918 \\
Adposition/noun order & \textsc{adp} $\xleftarrow{\text{case}}$ \textsc{noun} & 97.6\% & 0.975 & 98.1\% & 0.980 \\
Prefixing & Prefixes $\geq$ 50\% & 87.3\% & 0.808 & 85.6\% & 0.798 \\
Suffixing & Suffixes $>$ 50\% & \multicolumn{4}{c}{Identical to \featlabel{Prefix} in this condition} \\
        \bottomrule
    \end{tabular}
    \caption{Projected properties. The word classes \textsc{adj*} and \textsc{num*} are narrower versions of the corresponding UD word classes, see the main text for details. Accuracy and F$_1$ values are with respect to URIEL values from WALS and Ethnologue. \textbf{Exclusive} counts only languages where URIEL codes exactly one of a mutually exclusive set of options as true, \textbf{Non-exclusive} uses all available data. \textbf{Language} gives each ISO 639-3 language code equal weight, while \textbf{Family} gives each Glottolog family identifier equal weight.}
    \label{tab:projected-word-order}
\end{table}

At this point we should add that the classifications derived from projected data are never assumed to be correct in our evaluation. Instead, they are used as \emph{training} data in some of our classification experiments, while only URIEL is used as a gold standard for comparison. We do however use projected labels as a complement in our error analysis in \Fref{sec:error-analysis}.

\section{Language representations}
\label{sec:languagereps}

In order to capture different types of linguistic structure, we use a number of different neural models for creating language representations.\footnote{For languages with multiple Bible translations, we learn one representation per translation (doculect). The exception is the ASJP-based model and the language representations from previous work, which are all on the (ISO 639-3) language level. For simplicity, we use \emph{language representation} for both levels of granularity.} The model types are chosen to maximize the diversity of the learned representations, while requiring only the available data described in Sections \ref{sec:data} and \ref{sec:projection}.

We use the following models, which generate the language representations whose labels are in bold:
\begin{itemize}
    \itemsep0em 
    \item Word-based language model with multilingual word embeddings (\replabel{WordLM})
    \item Character-based language model (\replabel{CharLM})
    \item Morphological reinflection of noun paradigms (\replabel{Reinflect-Noun}) or verb paradigms (\replabel{Reinflect-Verb})
    \item Word form encoder from characters of a word form to the multilingual word embedding space (\replabel{Encoder})
    \item Neural machine translation models: many-to-English (\replabel{NMTx2eng}) and English-to-many (ŕeplabel{NMTeng2x})
    \item Baseline representations from pairwise lexical similarity (\replabel{Lexical} and \replabel{ASJP})
\end{itemize}
The models will be detailed in the following subsections.

\subsection{Word-level language model}
\label{sec:wordbased}

We train a language model to predict the location of the following word in the multilingual word embedding space. This consists of a simple left-to-right LSTM conditioned on the preceding word and a language embedding, whose output is projected through a fully connected layer to the multilingual word embedding space. As loss function, we use the cosine distance between the predicted word embedding and the actual word at that position in training data. This allows us to efficiently train the model with a vocabulary size of 18 million word types.

Only the LSTM parameters, the fully connected layer following it, and the language representations are updated during training. The word embeddings are fixed. Sentences of all languages are mixed, and presented in random order. In the experiments, we use 512-dimensional LSTM with 100-dimensional language embedddings. For the regularization, we use a dropout layer with probability 0.3 between the LSTM and the hidden layer.

Since semantic information is encoded in a language-independent way by the multilingual word embeddings, our intention with this model is for the LSTM to learn a language-agnostic model of semantic coherence, while relying on the language representations to decide how to order the information---that is, the syntax of each language.  We refer the representations obtained from this model as \replabel{WordLM}.

\subsection{Character-based language model}
\label{sec:characterbased}

We train a single LSTM language model over the characters making up each sentence in all languages. The model is conditioned at each time step only on the preceding character and a language embedding. The character embeddings are shared between languages. Sentences from all languages are mixed, and presented in random order. All parameters of the model are learned from scratch during training.

Ideally, we would want to train this model using an accurate transcription in e.g.\ the International Phonetic Alphabet (IPA), but the Bible corpus is generally only available in the standard orthography (or orthographies) of each language. Since a number of very different writing systems are used, it is not possible to directly use the raw text. To approximate a phonemic transcription, we use standard transliteration\footnote{Using the transliteration tables from the Text::Unidecode library of Sean Burke.} into Latin script, followed by a few rules for phonemes generally represented by multi-grapheme sequences across Latin-based orthographies (e.g.\ \textit{sh} $\rightarrow$ \textesh), as well as merging some vowels and voicing distinctions to reduce the size of the inventory. If accurate multilingual grapheme-to-phoneme (G2P) systems become available that cover most of languages in the Bible corpus, that would of course be a much preferred solution since our approximations are not valid for all languages and orthographies.

This is roughly equivalent to the model of \citet{Ostling2017multilm}, except that we use a pseudo-normalized Latin orthography rather than native writing systems. We refer to the model as \replabel{CharLM}. We use a 128-dimensional LSTM, 100-dimensional character embeddings and 100-dimensional language representations. We also use a dropout layer with probability 0.3 between the LSTM and the dense layer for regularization.

\subsection{Multilingual reinflection model}
\label{sec:reinflection}

We train an LSTM-based sequence-to-sequence model with attention to predict one form in an inflectional paradigm given another form. In spirit, this is similar to the reinflection task of \citet{Cotterell2016sigmorphon}, except that we do not have access to accurate annotations of morphological features. Instead we simply pick random target forms without providing the model any further information. This model is implemented with OpenNMT \citep{Klein2017opennmt} using default hyperparameters. We train two sets of language representations: (i) Using only noun paradigms (\replabel{Reinflect-Noun}), (ii) Using only verb paradigms (\replabel{Reinflect-Verb}).\footnote{As a sanity check, we have sampled from the model and as expected the k-best list of translations generally contains correct (but arbitrary) inflections of the lemma that the source form belongs to.} The target language is represented by a special token for each language, whose embedding becomes the language embedding for that language.

The model has direct access to the source form through the attention mechanism, and our intention is that it will learn to copy the lexical root of the source form to the target, needing only to learn which transformations to apply (e.g.\ removal and addition of affixes), and not to memorize the vocabularies of all languages. We expect the language representations to encode the necessary morphological information to perform this transformation. This is similar to the use of morphological inflection for fine-tuning language representations in \citet{Bjerva2018fromphonology}, except that we rely only on cross-lingual supervision and are thus able to directly train the model for the whole Bible corpus.

\subsection{Word encoder model}
\label{sec:encoder}

The reinflection model described in the previous section is only concerned with predicting some other member of the  same inflectional paradigm, without considering the properties of that form. It is therefore not possible for the model to connect a certain form with, say, number marking on nouns or tense marking on verbs. For this reason, we also train a model to encode word forms represented as transliterated character sequences into the multilingual word embedding space from \Fref{sec:embedding-projection}. This model consists of a $2 \times 128$-dimensional BiLSTM encoder over a character sequence, followed by an attention layer and a fully connected layer. We use cosine distance loss, as in the multilingual language model from \Fref{sec:wordbased}. The target language is represented by a special token for each language, whose embedding becomes the language embedding for that language.

Our aim with this model is to capture not only general tendencies of inflectional morphology, but also the presence and location of specific markers (such as case suffixes, or number prefixes). We refer the representations obtained from this model as \replabel{Encoder}.

\subsection{Machine translation models}
\label{sec:nmt}

Inspired by \citet{Malaviya2017learning}, we train a many-to-English (\replabel{NMTx2eng}) and an English-to-many (ŕeplabel{NMTeng2x}) neural machine translation system. These are implemented in OpenNMT \citep{Klein2017opennmt}, using 512-dimensional LSTM models with a common subword vocabulary on the transliterated and normalized data described above in \Fref{sec:characterbased}. For the many-to-English model, the source language is encoded using a unique token per language, while for the English-to-many model it is the target language that is encoded by a unique token. The embeddings of these tokens are used as language representations.

\subsection{Lexical similarity}
\label{sec:asjp}

For comparison purposes, we include two non-neural baselines which  contain only \emph{lexical} information about languages. The first is derived from the ASJP lexical database \citep{asjp}, which contains 40-item word lists of core vocabulary for a large number of languages. A total of 1012 languages (unique ISO 639-3 codes) occur in the Bible corpus and have sufficiently complete (at least 30 items) word lists in ASJP. We follow \citet[p.\ 171]{bakker2009addingtypology} in measuring the distance between two languages by taking the mean normalized Levenshtein distance between same-concept word forms, divided by the mean normalized Levenshtein distance between different-concept word forms.\footnote{For consistency with \citet{bakker2009addingtypology}, we normalize by dividing by $\max(|s_1|,|s_2|)$.} If multiple varieties of the same (ISO 639-3) language are present in ASJP, the union of word forms over all variteties is used. We compute a $1012 \times 1012$ pairwise distance matrix, which we reduce to 100 dimensions using
truncated SVD as implemented by \citet{scikit-learn}.\footnote{We also attempted to use UMAP \citep{mcinnes2018umap-software}, but found the structure of the resulting vectors to lead to instability during classifier training.}
We refer to this set of language representations as \replabel{ASJP}.

\section{Experiments}
\label{sec:experiments}

As set out in the introduction, we are interested in finding out to what extent we can control the type of information captured by language representations, and whether language embeddings from neural models make human-like typological generalizations. We do this by answering, for a large number of typological features $f$, how well a given set of language representations $L$ capture $f$. Specifically, we find the extent to which $f$ can be predicted from $L$ alone using a logistic regression classifier. For ease of analysis, we train a binary logistic regression classifier for each feature with equal weights for the positive and negative class. This avoids biasing classifiers according to the data label distribution, which allows easier comparison between different subsets of the data, with different label distributions. In addition, our sampling procedure (described further below) gives equal weight to language families, regardless of how many members they contain.

\subsection{Evaluation data}
\label{sec:evaluationdata}

The typological features used in this study are derived from two types of sources: traditional typological databases (following e.g.\ \citealt{Malaviya2017learning}), as well as a novel dataset consisting of word order features obtained from annotation projection in the Bible corpus.

\subsubsection{Typological databases}
We use the URIEL typological database \citep{Littel2017uriel}, specifically the features  derived from the World Atlas of Language Structures (WALS, \citealt{wals}) and Ethnologue \citep{Ethnologue}.
Features from these sources are used as gold standard labels for the evaluation.
Note that the binarization of features in URIEL requires some simplification to the (already simplified) coding in the original data source. Features representing several mutually contradictory values may simultaneously be true. For instance, Irish is coded in URIEL as tending towards suffixing morphology, but also tending towards prefixing  (it is coded as ``Equal prefixing and suffixing'' by \citet{wals-26}), while German according to URIEL has both object after verbs and object before verbs (it is coded as ``No dominant order'' by \citet{wals-83}). We resolve this by keeping only those instances in the data where exactly \emph{one} of a set of mutually incompatible variables is true.

\subsubsection{Projected features}
Five types of projected word order statistics described in \Fref{sec:wordorderstatistics} (object/verb order, subject/verb order, adjective/noun order, numeral/noun order, adposition/noun order)
are used as training data for the classifiers, but never as gold standard labels for evaluation. This data has the advantage of being available for all languages in the Bible corpus, which allows more languages to be used for training than if we would restrict ourselves to the languages present in URIEL for the given feature. In addition, the morphological feature indicating whether prefixing or suffixing morphology dominates is used.

\subsection{Cross-validated classification}
\label{sec:crossvalidation}

Our basic measure of whether a set of language representations encode a specific typological feature is cross-validated classification performance, measured using F$_1$ score (the mean of the F$_1$ of the positive and negative classes). As described in \Fref{sec:framework}, we use constrained leave-one-out cross-validation, taking care to exclude languages from the training fold that could be suspected to be non-independent of the evaluated language. All languages with gold standard labels available are classified, and the results are weighted in order to give either languages (defined according to ISO 639-3 codes) or language families (defined according to Glottolog family identifiers) equal weight. We consider family-weighted F$_1$ score to be the single most useful measure of classifier success, and this is what we report unless otherwise specified.

The uncertainty is estimated by Monte Carlo sampling, where 401 samples are drawn such that only one language from each family is chosen. As a dummy baseline, we train classifiers using the same parameters and data but with randomly shuffled target labels. This establishes a baseline range of F$_1$ and accuracy values that would be expected from a classifier that has not learned to predict the given feature at all.\footnote{We find that this baseline chance level agrees well with a binomial(0.5) model, as expected. Computing this baseline empirically rather than relying on a theoretical model helped us to diagnose an issue with insufficient regularization.} The non-baseline classifier variance across Monte Carlo samples is due to different training folds being chosen each sample. When a single classification is extracted, the type value across all samples is used.

If less than 50 language families are represented in the evaluation set for a particular feature, we skip evaluating it due to data sparsity.

\section{Results}
\label{sec:results}

We will now describe the results of our evaluations for our own models (see \Fref{sec:languagereps}) as well as of two previous studies. From \citet{Malaviya2017learning} we use two sets of language representations derived from the same model: \replabel{MTVec} (language embeddings) and \replabel{MTCell} (averaged LSTM cell states). From \citet{Ostling2017multilm} we use the concatenated embeddings that were fed into the three LSTM layers, here labelled \replabel{Ö\&T}. Some other authors have investigated language representations for smaller sets of languages, but our evaluation setup is unsuitable for samples much smaller than a thousand languages.

In the figures below, we present the mean family-weighted F$_1$ for each set of language representations, for each feature of interest. Language representations are grouped in five groups that are visually distinguished in the figures:
\begin{enumerate}
    \item Lexical baselines: \replabel{ASJP} and \replabel{Lexical}. These should, by design, not encode any structural features of language.
    \item Neural Machine Translation (NMT): our \replabel{NMTx2eng} and \replabel{NMTeng2x} models, as well as \replabel{MTCell} and \replabel{MTVec} from \citet{Malaviya2017learning}.
    \item Character-level language models: our \replabel{CharLM} and the previously published \replabel{Ö\&T} \citep{Ostling2017multilm}.
    \item Word-level language model: our \replabel{WordLM}.
    \item Word form models: our \replabel{Reinflect-Noun}, \replabel{Reinflect-Verb} and \replabel{Encoder}.
\end{enumerate}
Each figure has a dotted line indicating the 99th percentile of the shuffled-label baselines. This should be seen as a very rough baseline indicator, since we do not have a good way of modeling the complex distribution of classification results obtained from the (hypothetical) set of all possible language representations that do not encode typological features, given our sampling distribution of training languages. Language representations derived from lexical similarity exceed this baseline in two cases, though only by a small amount, so it likely represents an under-estimation of the actual baseline distribution. We do not interpret results exceeding this baseline as definite confirmations of typological features being encoded in the given language representations.

Some figures also have a dashed line, indicating the mean F$_1$ of projected labels with respect to the gold standard in URIEL. These correspond to the rightmost column in \Fref{tab:projected-word-order}. We include the projection performance because it represents what can be done using hand-crafted methods on the same parallel text data as we have used for creating the language representations. Reaching this level indicates that the classifier has likely become about as good as can be expected given the underlying data.

Note that some representations (\replabel{ASJP}, \replabel{MTCell}, \replabel{MTVec}, \replabel{Ö\&T}) are based on other data or other versions of the Bible corpus with a different set of languages, and thus have slightly different baselines. We have computed the baselines individually for each set of language representations to confirm that our conclusions hold, but choose not to represent this in the figures for readability. The dotted and dashed lines in the figures are generated from the version of the Bible corpus used by us.

We wish to emphasize that if a set of language representations encode a typological feature in a useful way, given the hundreds of data points we use for training, we expect the classifier to be highly accurate. In contrast, with our evaluation setup we expect classifiers to perform (approximately) randomly if there are no relevant typological features encoded in the language representations used to train them. Since the relevant differences in classification accuracy are very large, we present the main results as bar plots, complemented by exact numbers in the text only when we deem relevant. Differences between poorly performing classifiers are not relevant for our purposes, and we refrain from summarizing the complete data in a separate table. We should add that correlations between typological features somewhat complicate this binary distinction, but this is only relevant for the few language representations that actually seem to encode typological features, and those are analyzed in detail below.

\subsection{Word order features}

\begin{figure}[p]
    \centering
    \begin{subfigure}[b]{0.95\textwidth}
        \includegraphics[width=\textwidth]{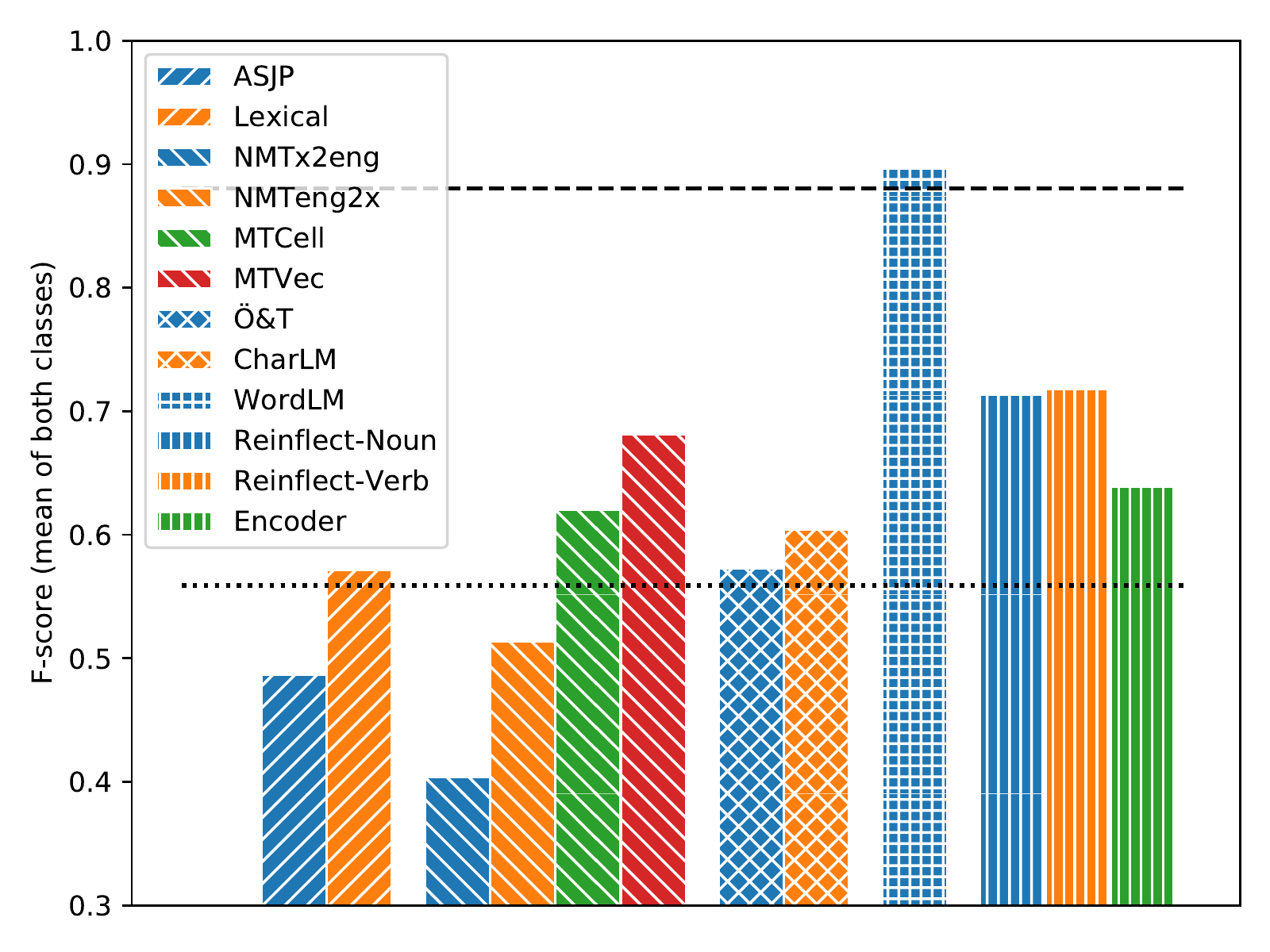}
        \caption{Order of object and verb, using gold standard labels for training.}
        \label{fig:ov_uriel}
    \end{subfigure}
    \vfill
    \begin{subfigure}[b]{0.95\textwidth}
        \includegraphics[width=\textwidth]{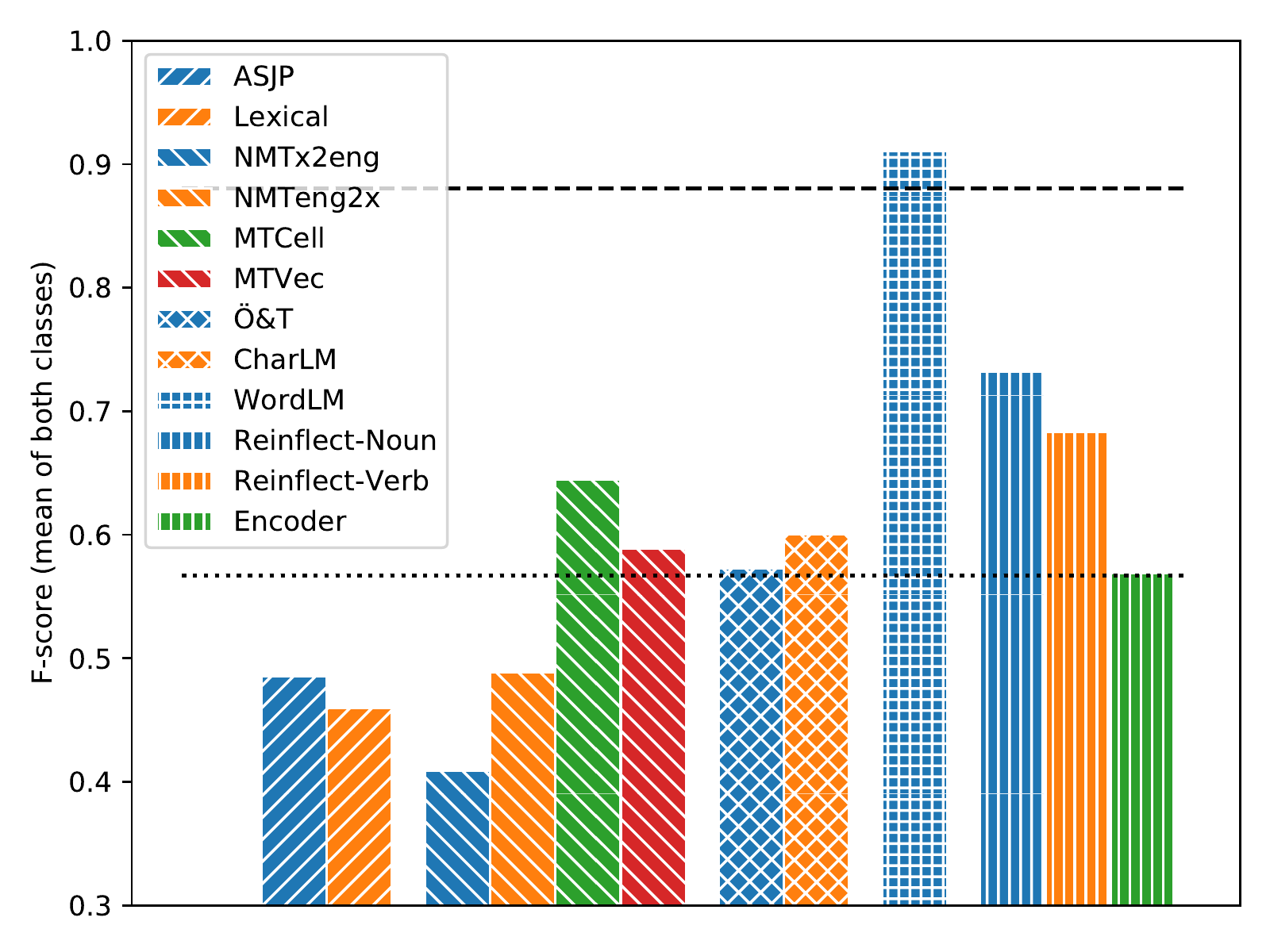}
        \caption{Order of object and verb, using projected labels for training.}
        \label{fig:ov_proj}
    \end{subfigure}

    \caption{Classification results for each set of language representations.}
    \label{fig:ov}
\end{figure}

We start by looking at \Fref{fig:ov_uriel}. The first thing to notice is that only the language representations from our word level language model (\replabel{WordLM}) reach an F$_1$ score comparable to (and even slightly above) that of the projection method. This indicates that only the word level language model has managed to capture the order of object and verb, at least in a way that is separable by a linear classifier. The lexical baselines (\replabel{ASJP} and \replabel{Lexical}) encode lexical similarity between languages, and so are strongly correlated with word order properties within related languages or languages in contact. As intended, our evaluation setup prevents these models from learning to identify even a clear and evenly distributed feature like the order of object and verb. Character-level language models (\replabel{CharLM} and \replabel{Ö\&T}) do not seem to encode word order properties, which indicates that they have not learned representations at the syntactic level. This is not surprising, since both models are relatively small and unlikely to learn enough vocabulary to generate to the level of syntax.

\begin{figure}[p]
    \centering
    \begin{subfigure}[b]{0.95\textwidth}
        \includegraphics[width=\textwidth]{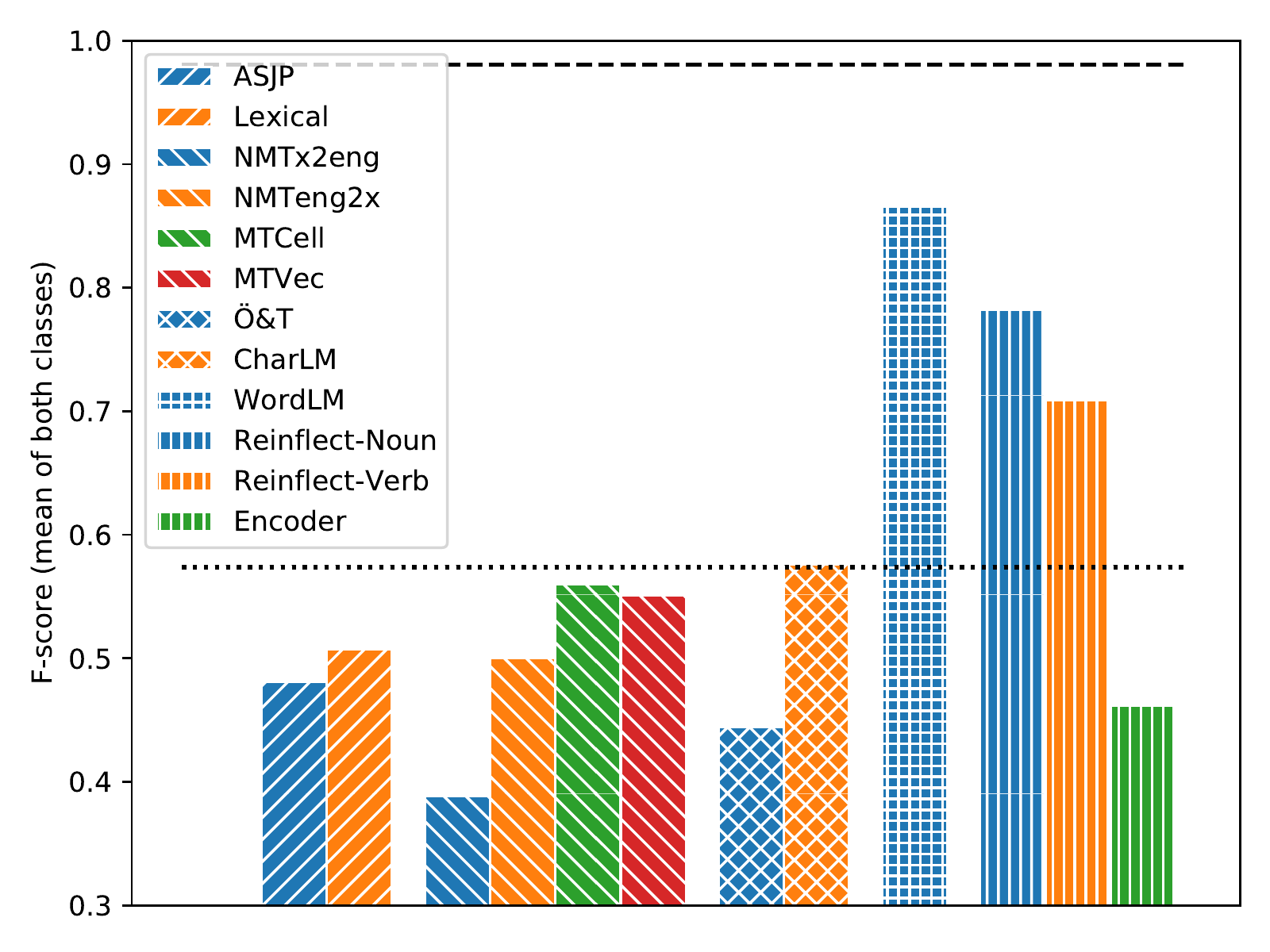}
        \caption{Prepositions vs postpositions, using gold standard labels for training.}
        \label{fig:adp_uriel}
    \end{subfigure}
    \vfill
    \begin{subfigure}[b]{0.95\textwidth}
        \includegraphics[width=\textwidth]{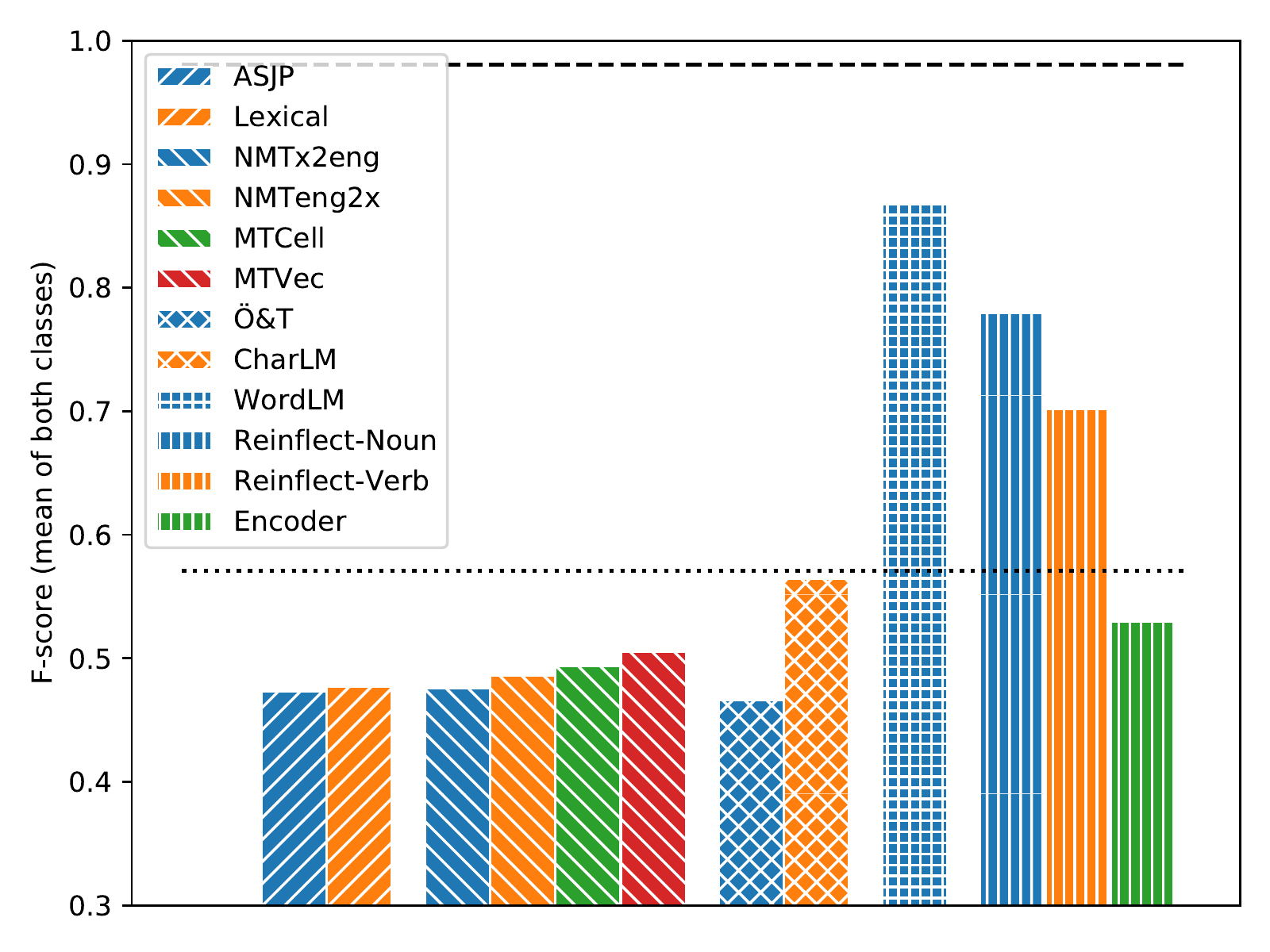}
        \caption{Prepositions vs postpositions, using projected labels for training.}
        \label{fig:adp_proj}
    \end{subfigure}

    \caption{Classification results for each set of language representations.}
    \label{fig:adp}
\end{figure}

The word form models, in particular the reinflection models (\replabel{Reinflect-Noun} and \replabel{Reinflect-Verb}), obtain moderately high F$_1$ values of around 0.7. Yet it is obvious that these models do not have sufficient data to conclude what the order of object and verb are in a language, since their input consists entirely of automatically extracted inflectional paradigms. We therefore suspect that the relative success in predicting may be due to the classifiers learning to predict \emph{another} feature that correlates with the order of object and verb. To investigate whether this explanation is correct, we compute the corresponding F$_1$ scores for the classifier predictions with respect to each typological feature where we have data. In this case we find that classifications from both reinflection models are much better explained (\replabel{Reinflect-Noun}: +0.19 F$_1$, \replabel{Reinflect-Verb}: +0.05 F$_1$) by the affix position \citep{wals-26} feature.\footnote{Each pair of features has a unique set of overlapping languages, which we use in these comparisons in order to obtain comparable results. These F$_1$ differences from these head-to-head comparisons may not be equal to those obtained from using all available data for each feature, as we have presented in the figures.} In effect, the object/verb order labels we used for training were treated as noisy affix position labels, and the resulting classifier becomes much better at predicting affix position than object/verb order. An even clearer illustration of this can be found for the order of adposition and noun (see \Fref{fig:adp}), reflecting Greenberg's universal 27 \citep{Greenberg1963someuniversals} on the cross-linguistic association of prepositions with prefixing morphology, and postpositions with suffixing.

There has been a long-lasting debate on whether observed correlations between typological features are due to universal constraints on language, or simply due to genealogical and/or areal relations biasing the statistics \citep[e.g.][]{dunn2011lineage}. We remain agnostic with regards to this question, but note that analyzing correlations between typological features is a challenging statistical problem. In this work we test all other features for which we have data, and mention which ones seem like plausible alternative explanations for a given classification result in terms of comparable or higher F$_1$ scores, without attempting to quantify their relative probability of the different explanations.

\begin{figure}[p]
    \centering
    \begin{subfigure}[b]{0.95\textwidth}
        \includegraphics[width=\textwidth]{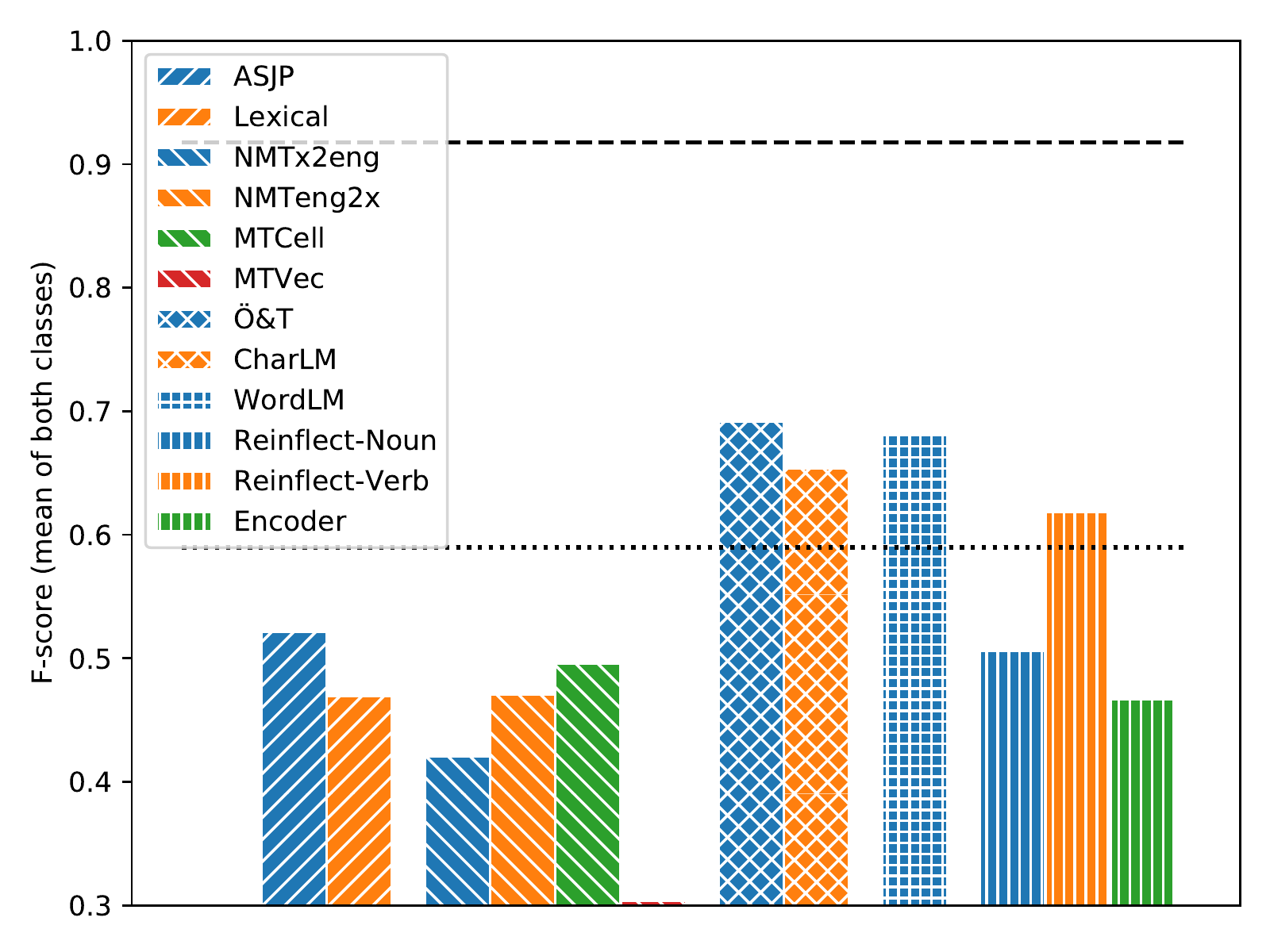}
        \caption{Order of numeral and noun, using gold standard labels for training.}
        \label{fig:numn_uriel}
    \end{subfigure}
    \vfill
    \begin{subfigure}[b]{0.95\textwidth}
        \includegraphics[width=\textwidth]{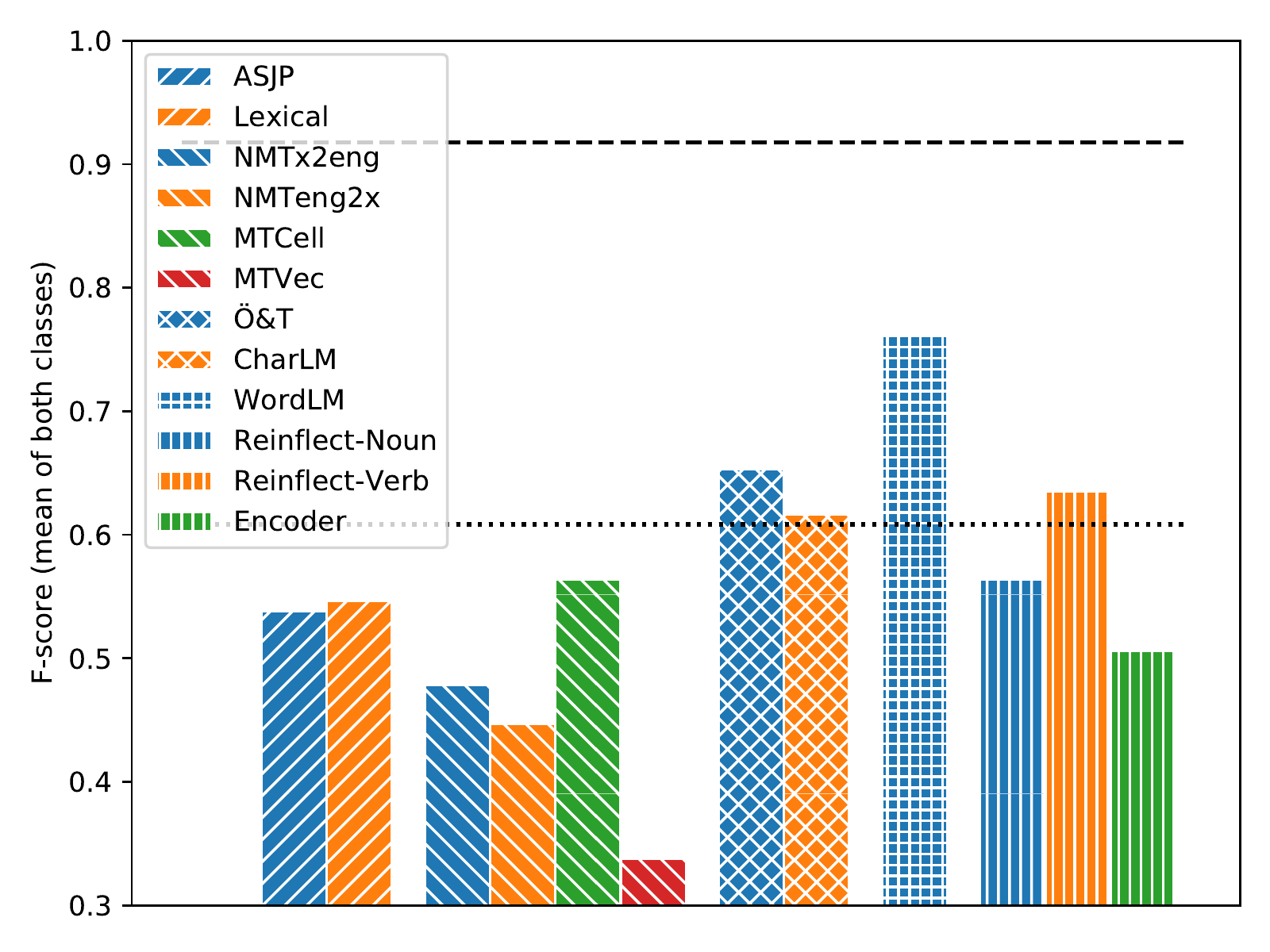}
        \caption{Order of numeral and noun, using projected labels for training.}
        \label{fig:numn_proj}
    \end{subfigure}

    \caption{Classification results for each set of language representations.}
    \label{fig:numn}
\end{figure}

\begin{figure}[p]
    \centering
    \begin{subfigure}[b]{0.95\textwidth}
        \includegraphics[width=\textwidth]{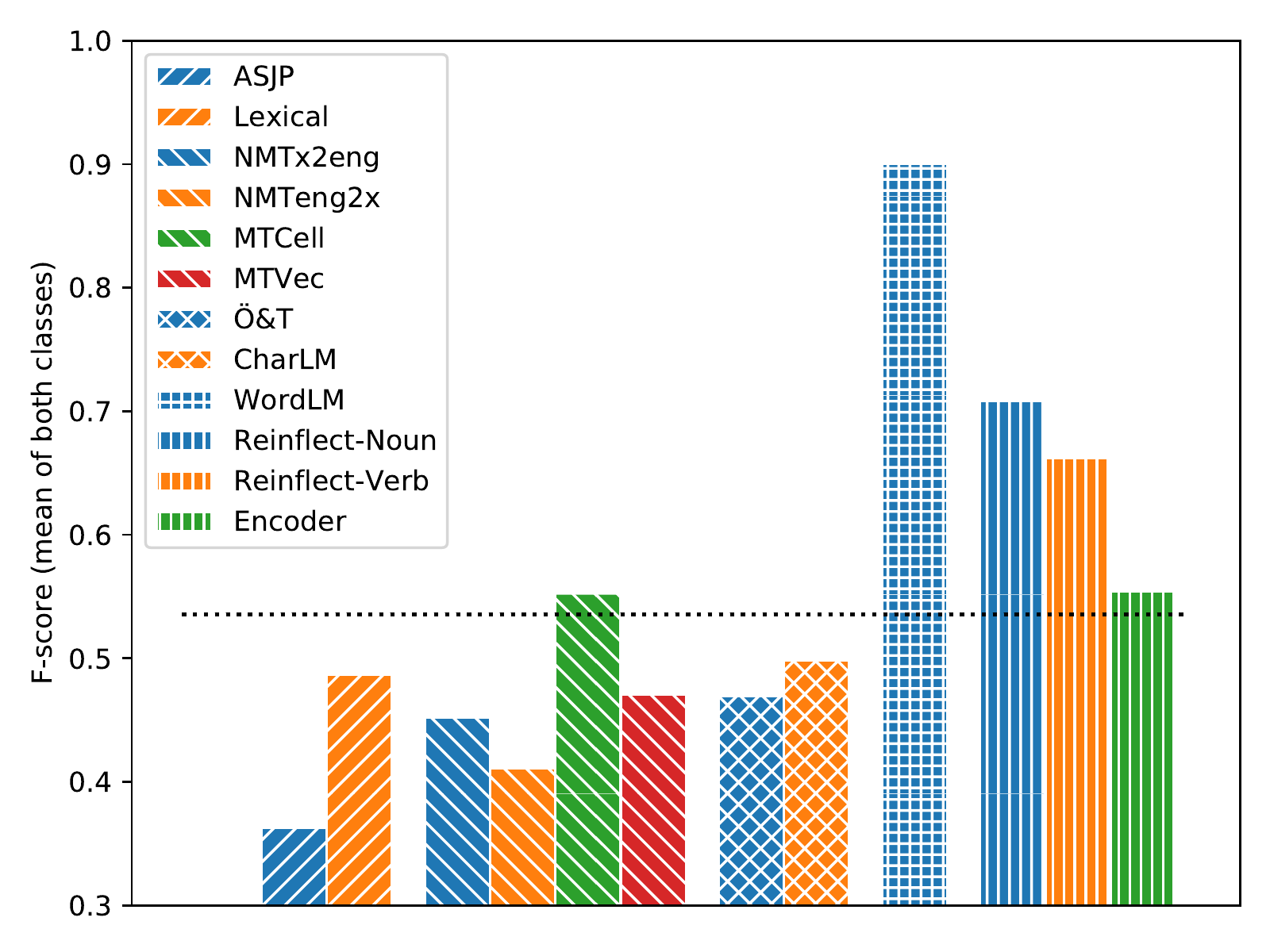}
        \caption{Order of possessor and noun, using gold standard labels for training.}
        \label{fig:possn_uriel}
    \end{subfigure}
    \vfill
    \begin{subfigure}[b]{0.95\textwidth}
        \includegraphics[width=\textwidth]{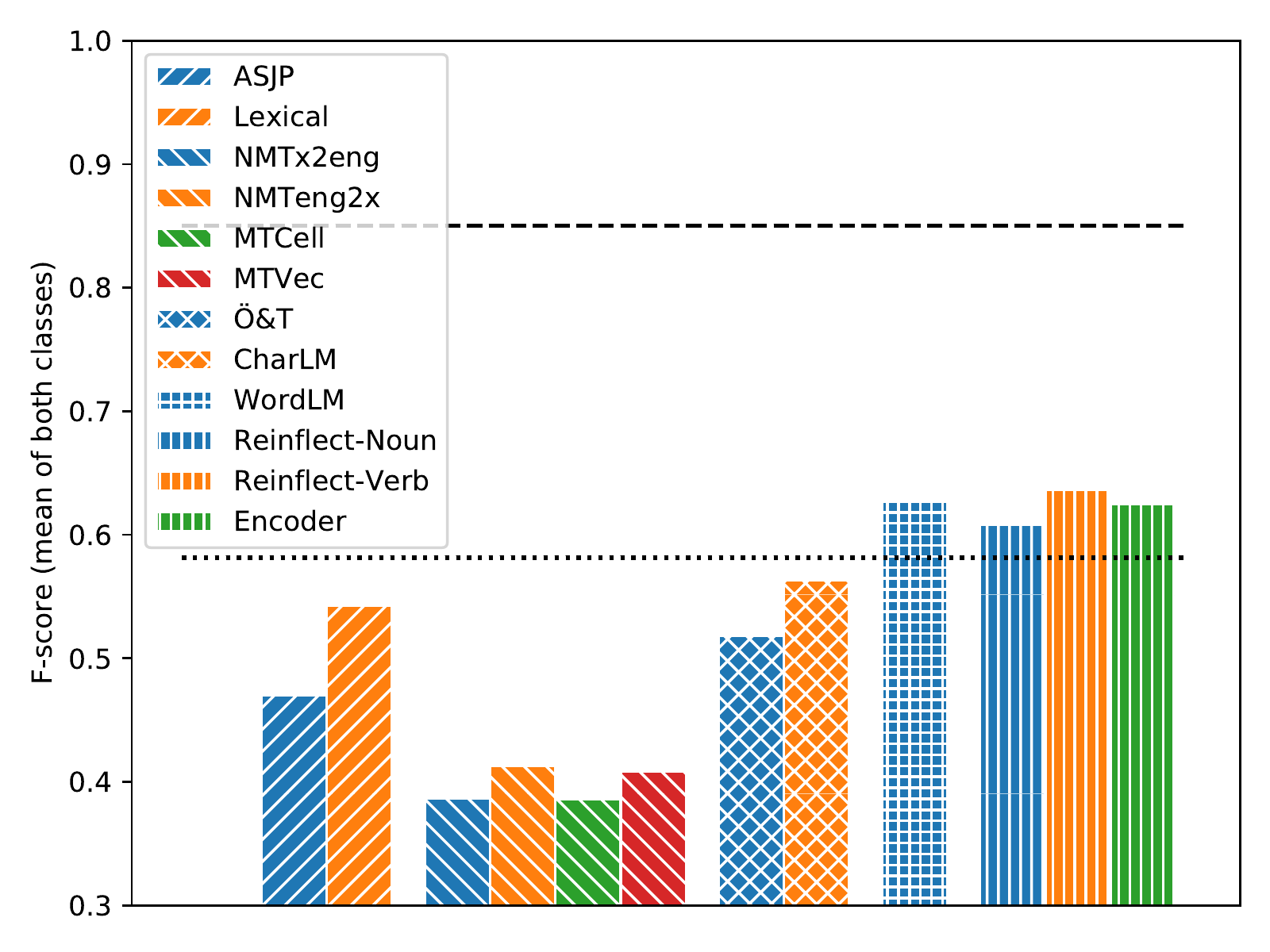}
        \caption{Order of adjective and noun, using gold standard labels for training. Version with projected labels is omitted, but very similar.}
        \label{fig:adjn_uriel}
    \end{subfigure}

    \caption{Classification results for each set of language representations.}
    \label{fig:possn_adjn}
\end{figure}

\begin{figure}[p]
    \centering
    \begin{subfigure}[b]{0.95\textwidth}
        \includegraphics[width=\textwidth]{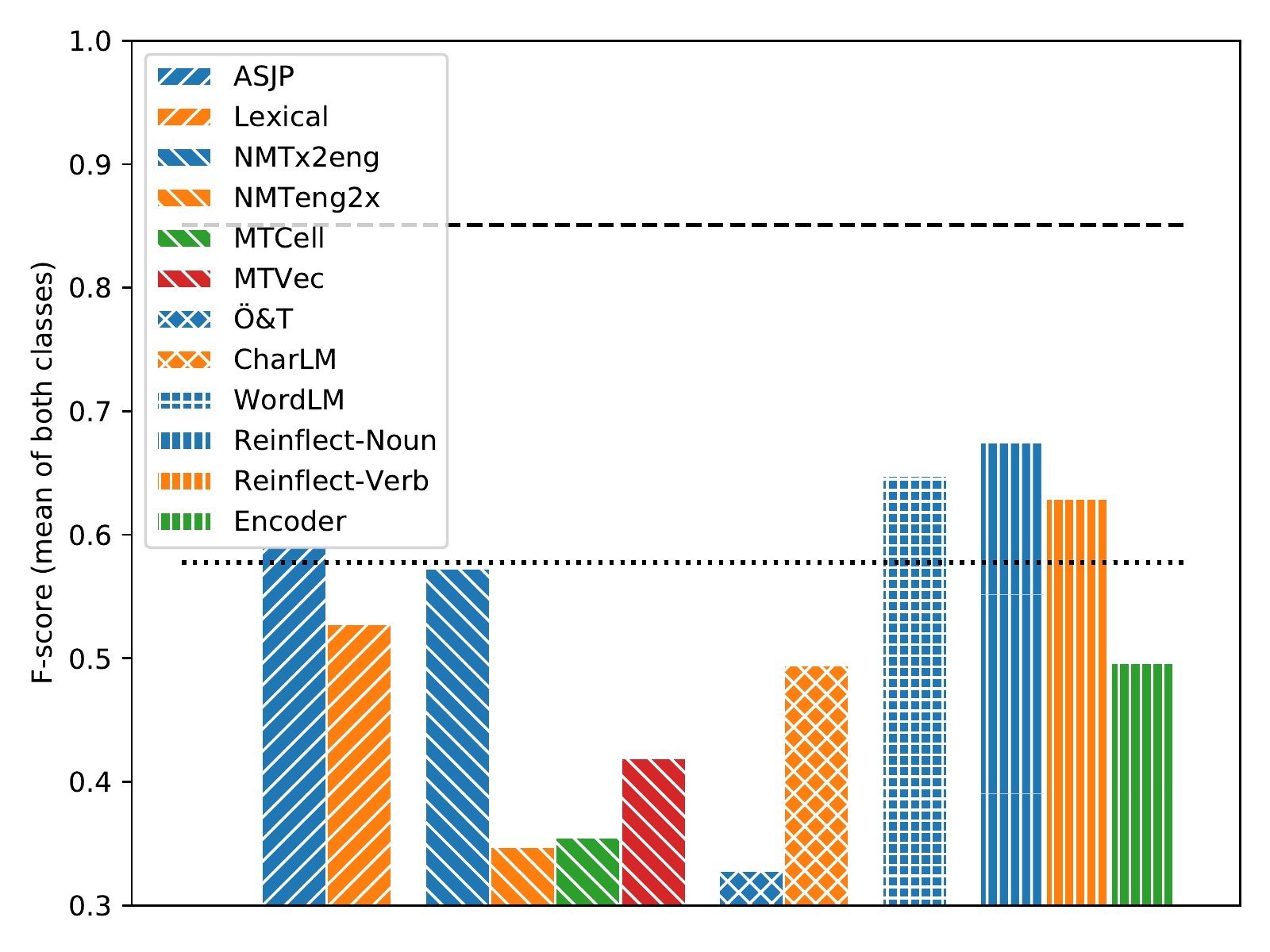}
        \caption{Order of relative clause and noun, using gold standard labels for training.}
        \label{fig:reln_uriel}
    \end{subfigure}
    \vfill
    \begin{subfigure}[b]{0.95\textwidth}
        \includegraphics[width=\textwidth]{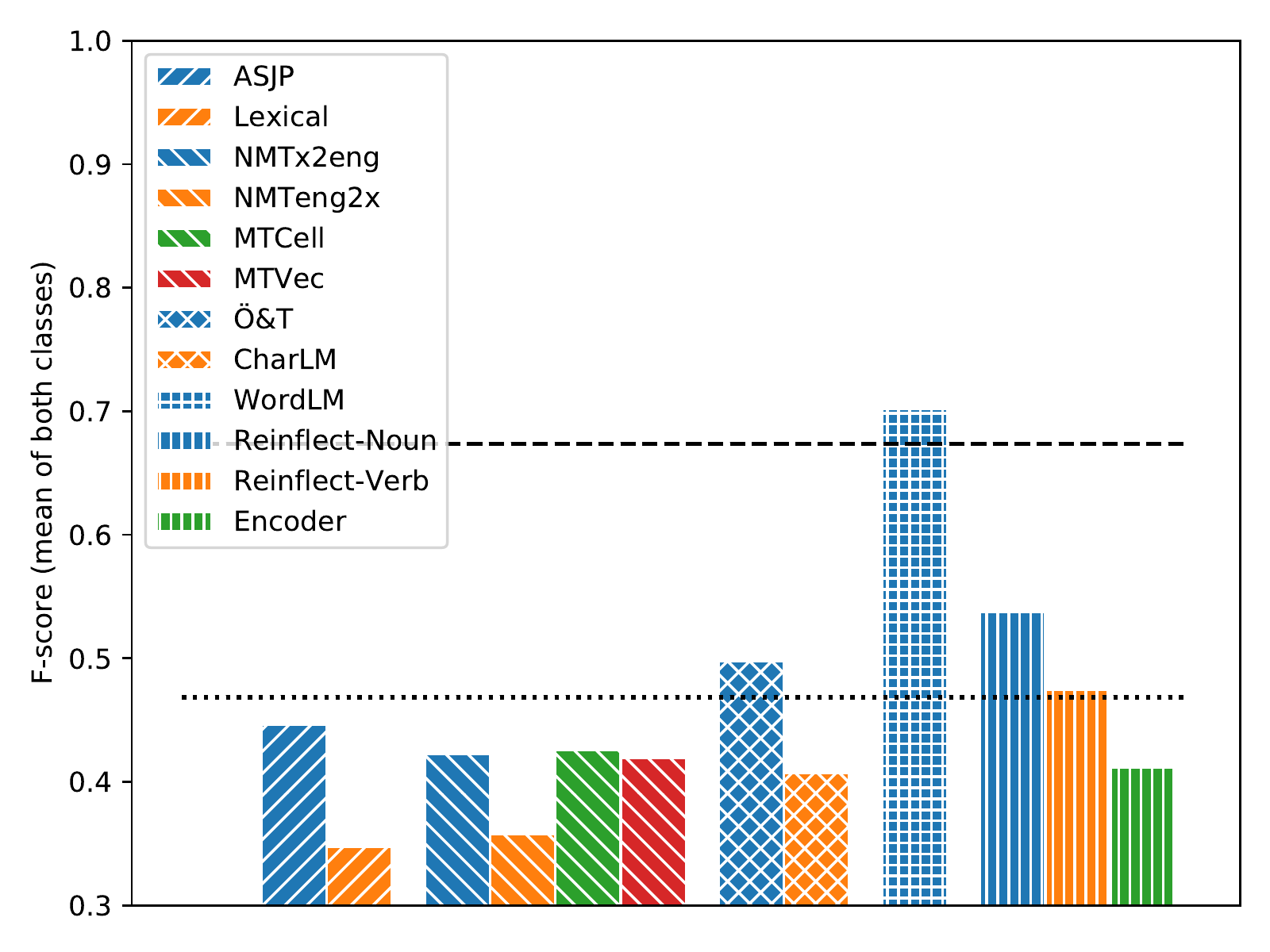}
        \caption{Order of subject and verb, using gold standard labels for training. Version with projected labels is omitted, but very similar.}
        \label{fig:sv_uriel}
    \end{subfigure}

    \caption{Classification results for each set of language representations.}
    \label{fig:reln_sv}
\end{figure}

\begin{figure}[p]
    \centering
    \begin{subfigure}[b]{0.95\textwidth}
        \includegraphics[width=\textwidth]{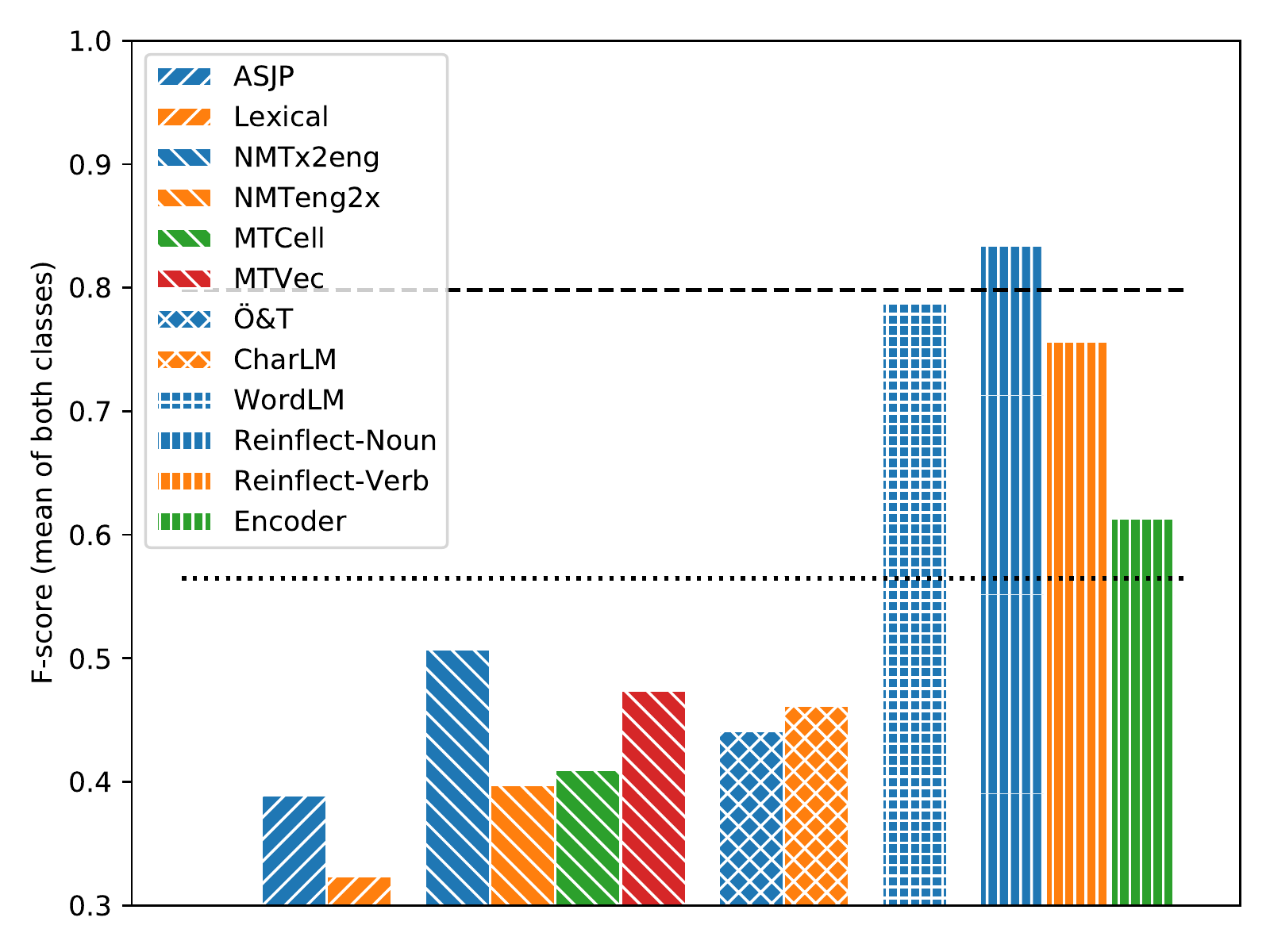}
        \caption{Prefixing or suffixing in inflectional morphology, using gold standard labels for training.}
        \label{fig:prefix_uriel}
    \end{subfigure}
    \vfill
    \begin{subfigure}[b]{0.95\textwidth}
        \includegraphics[width=\textwidth]{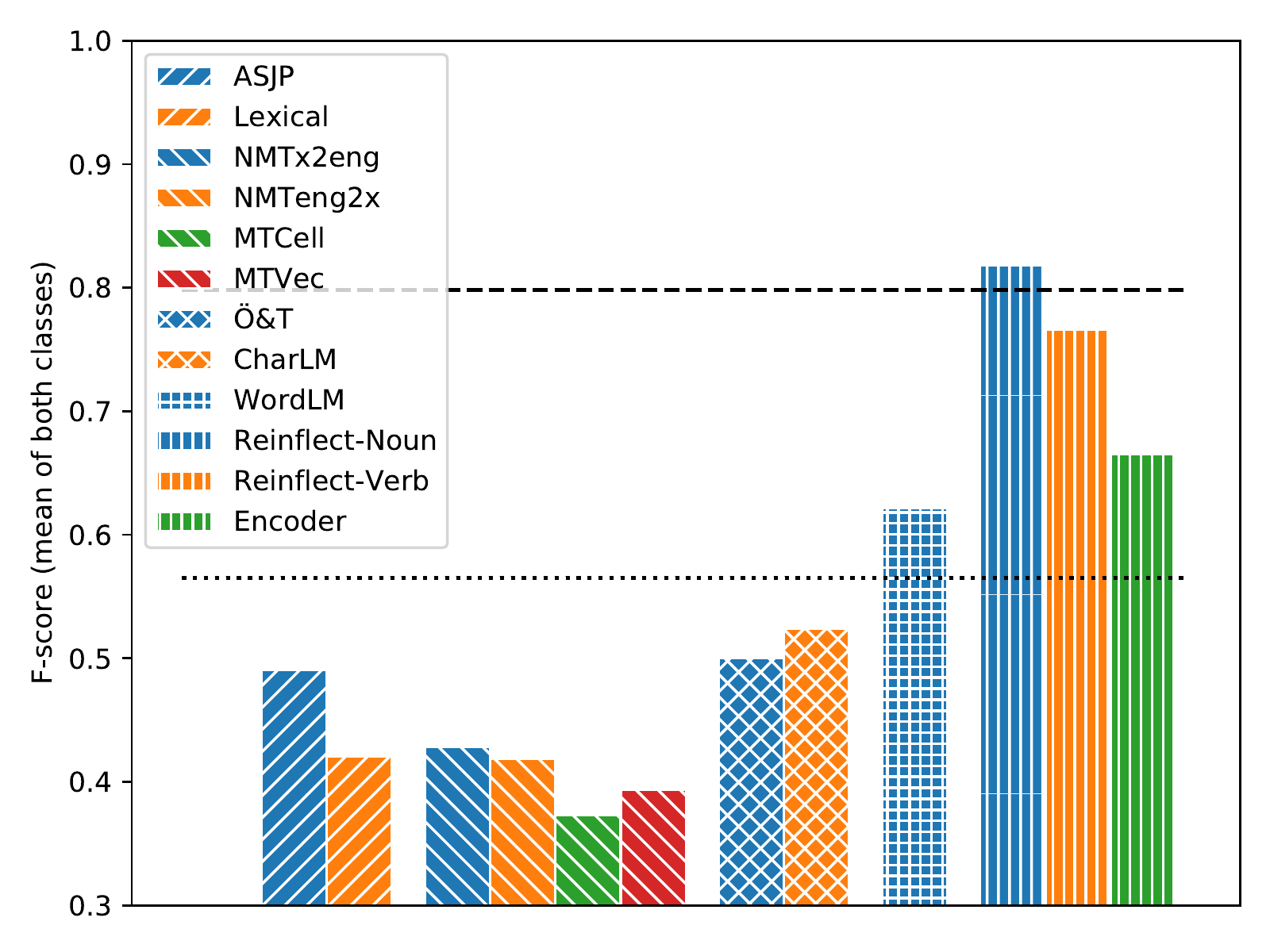}
        \caption{Prefixing or suffixing in inflectional morphology, using projected labels for training.}
        \label{fig:prefix_proj}
    \end{subfigure}
    \caption{Classification results for each set of language representations.}
    \label{fig:prefix}
\end{figure}

\begin{figure}[p]
    \centering
    \begin{subfigure}[b]{0.95\textwidth}
        \includegraphics[width=\textwidth]{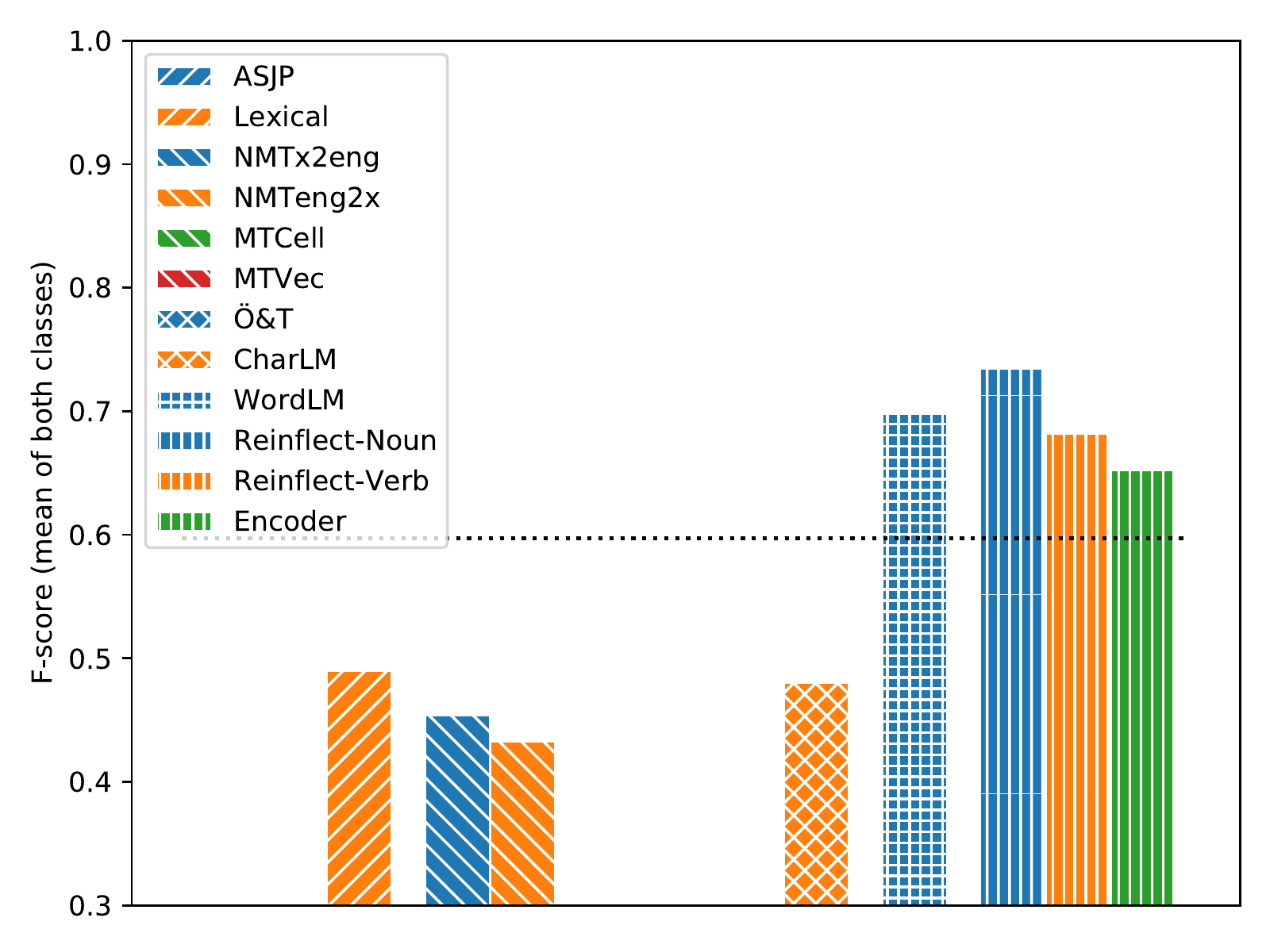}
        \caption{Negative prefix or suffix, using gold standard labels for training.}
        \label{fig:neg_prefix_uriel}
    \end{subfigure}
    \vfill
    \begin{subfigure}[b]{0.95\textwidth}
        \includegraphics[width=\textwidth]{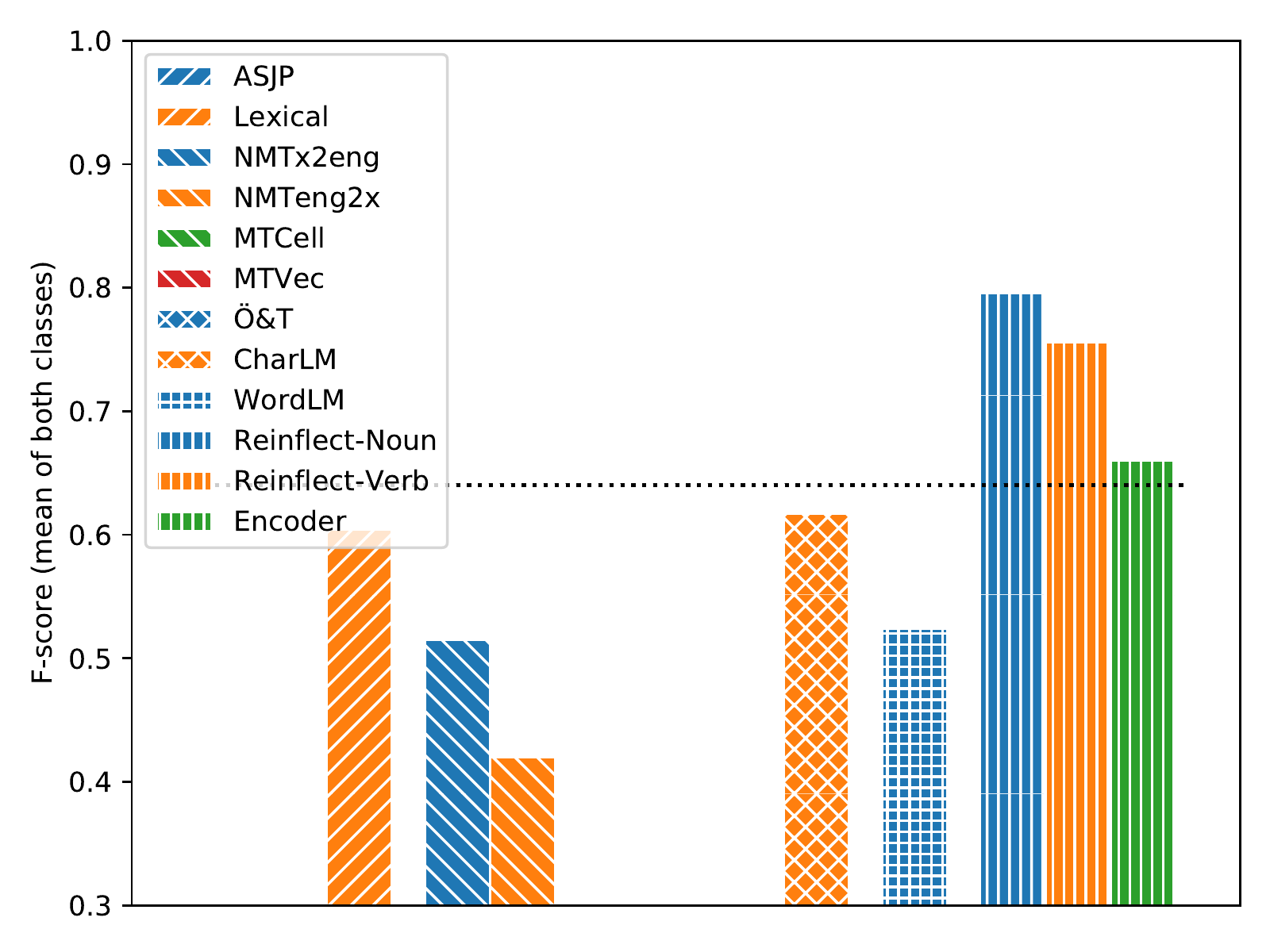}
        \caption{Possessive prefix or suffix, using gold standard labels for training.}
        \label{fig:poss_prefix_uriel}
    \end{subfigure}
    \caption{Classification results for each set of language representations. Note that some of the language representations contain too few languages in common with URIEL to be evaluated, the corresponding bars are omitted from the figures.}
    \label{fig:neg_poss_prefix}
\end{figure}

\begin{figure}[p]
    \centering
    \begin{subfigure}[b]{0.95\textwidth}
        \includegraphics[width=\textwidth]{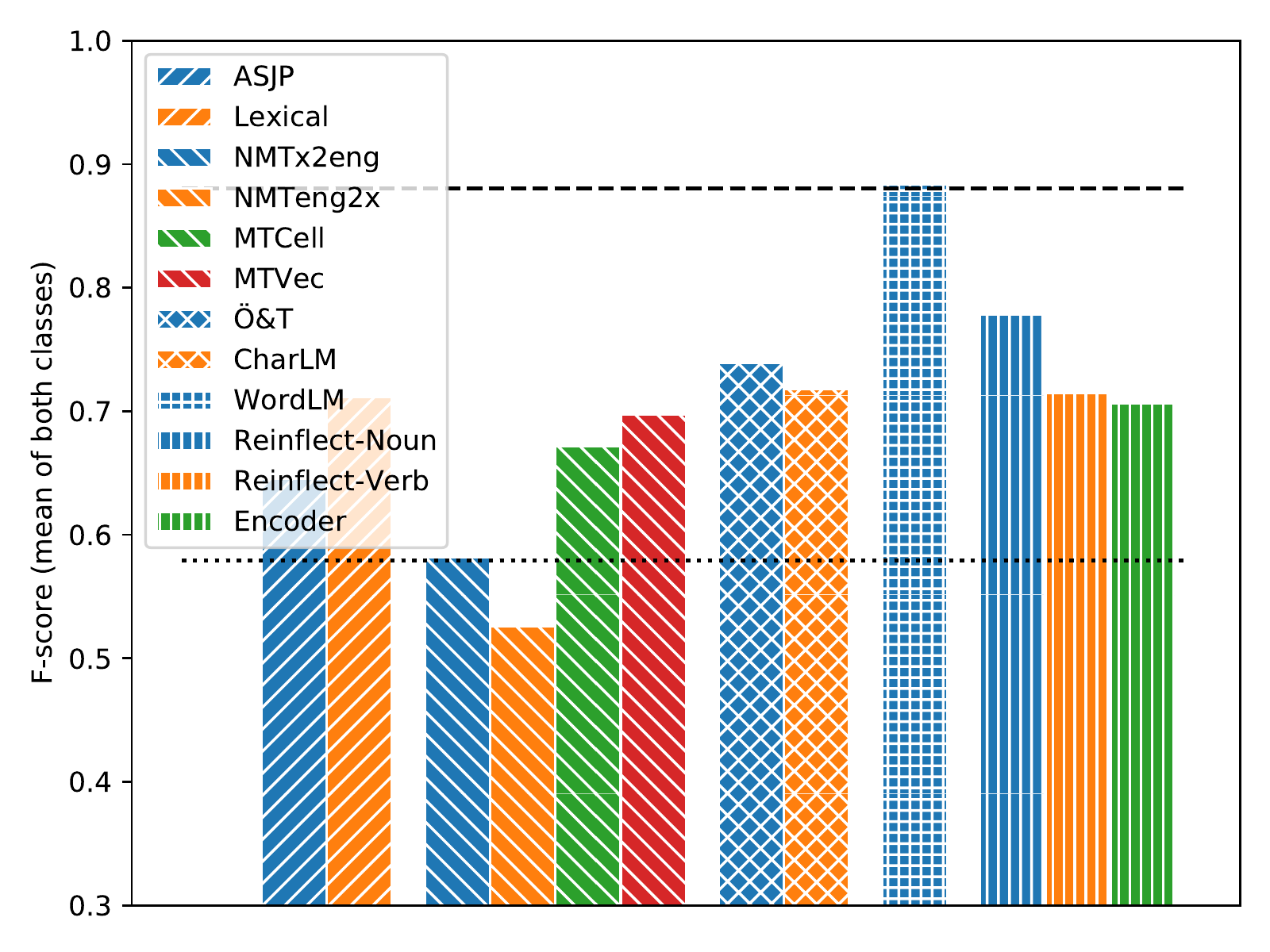}
        \caption{Order of object and verb, using gold standard labels for training and naive cross-validation.}
        \label{fig:naive_ov_uriel}
    \end{subfigure}
    \vfill
    \begin{subfigure}[b]{0.95\textwidth}
        \includegraphics[width=\textwidth]{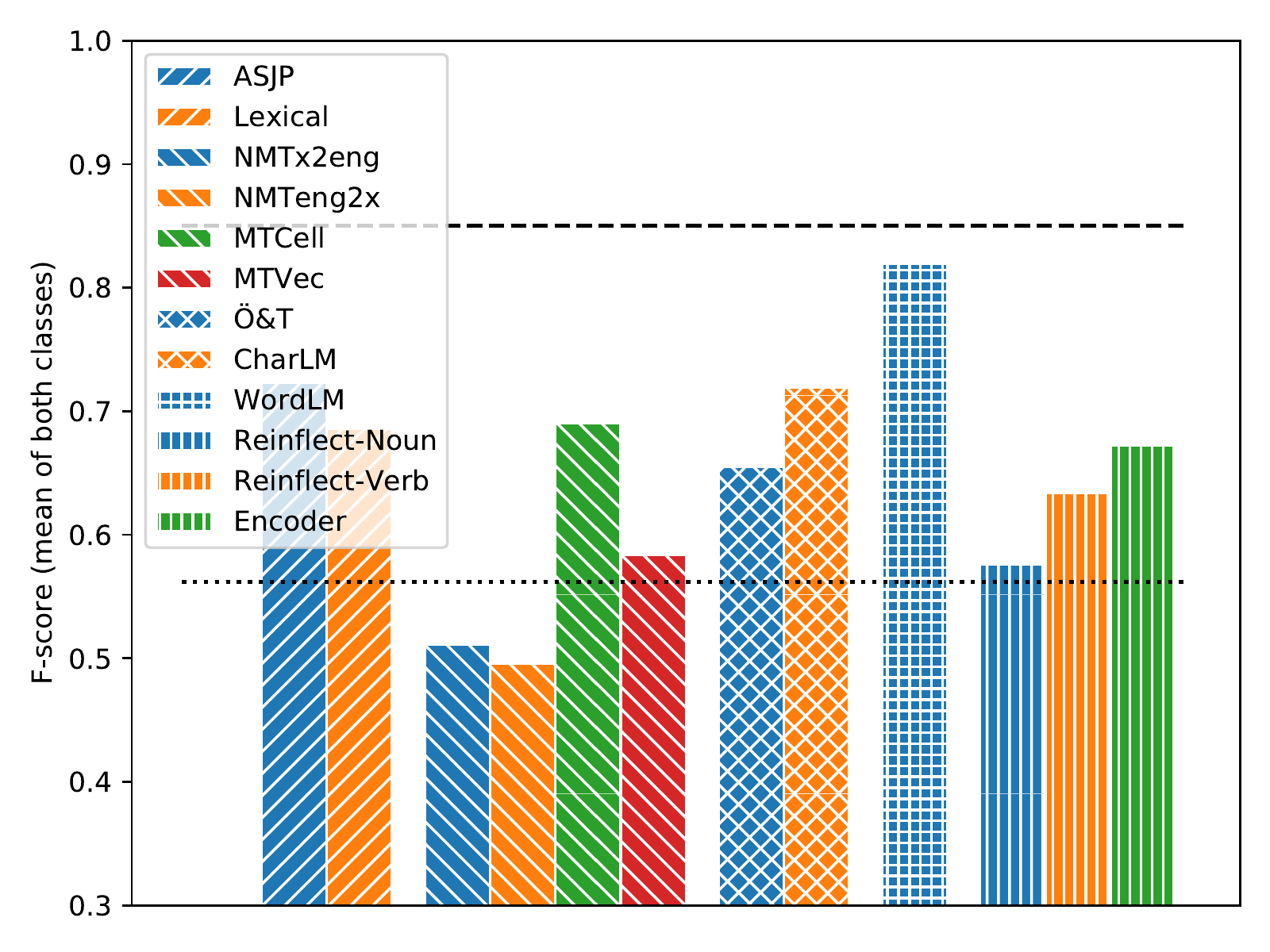}
        \caption{Order of adjective and noun, using gold standard labels for training and naive cross-validation.}
        \label{fig:naive_adjn_uriel}
    \end{subfigure}
    \caption{Classification results for each set of language representations, using naive cross-validation where languages related to the evaluated language are not excluded from the training fold. The point of this figure is to demonstrate how unsound evaluation methods give misleading results, see main text for details.}
    \label{fig:naive}
\end{figure}

To summarize, we observe clear detections of the following typological features related to word order:
\begin{itemize}
    \itemsep0em
    \item Order of object and verb, for the \replabel{WordLM} representations (\Fref{fig:ov}).
    \item Order of adposition and noun (prepositions/postpositions), for the \replabel{WordLM} representations (\Fref{fig:adp}).
    \item Order of numeral and noun, for the \replabel{WordLM} representations (\Fref{fig:numn}). Note that no representations obtained a mean F$_1$ above 0.7 when trained on URIEL data. As discussed in \Fref{sec:error-analysis}, this may be due to the much larger sample of languages with projected labels.
    \item Order of possessor and noun, for the \replabel{WordLM} representations (\Fref{fig:possn_uriel}). However, this result is about equally well explained (F$_1$ within 3 percentage points) by object and verb order, as well as adposition and noun order, so we consider this detection tentative.
\end{itemize}

In addition to the features presented in the figures, we also examined all other features in URIEL with sufficiently large samples for our evaluation method. The following features that relate to word order or the presence of certain categories of words were examined:
\begin{itemize}
    \itemsep0em 
    \item Order of demonstrative word and noun
    \item Order of relative clause and noun
    \item Order of subject and object
    \item Existence of a polar question word
\end{itemize}
None of the language representations yielded classifications with an F$_1$ above 0.7 for either of the above features.

It is interesting to note where we did \emph{not} see any clear indications of typological features encoded in the language representations. For at least some classical word order features, we see that there is sufficient information in the data to learn them, yet all models fail to do so.

The order of adjective and noun can be accurately projected (mean F$_1$ of 0.850, see \Fref{tab:projected-word-order}) but is not predictable with reasonable accuracy from even the \replabel{WordLM} representations (\Fref{fig:adjn_uriel}). This also applies to the order of relative clause and noun, with a projection F$_1$ of 0.851 but poor classification results (F$_1$: 0.648). Classifiers trained on relative/noun order become most proficient (F$_1$: 0.881) at classifying adposition/noun order.

As can be seen in \Fref{tab:projected-word-order}, the order of subject and verb is difficult to automatically extract through annotation projections in the data. The classification accuracy for the \replabel{WordLM} representations on this feature is somewhat better (0.702) than the projection result (0.673). For reasons discussed in \Fref{sec:error-analysis} below, we believe that this classifier has at least partly learned to identify subject/verb order.

Apart from \replabel{WordLM} and the \replabel{Reinflect} models, none of the representations reach a mean F$_1$ of 0.7 for any of the features under investigation.

\subsection{Morphological features}
\label{sec:morphological-features}

\Fref{fig:prefix} shows how well different language representations can be used to predict whether a language tends to use prefixes or suffixes (affixation type), according to the weighted affixation index of \citet{wals-26}. Languages classified as not using affixation, or with equal use of prefixes and suffixes, are excluded from the sample. The language representations best able to predict this feature is the \replabel{Reinflect-Noun}, followed by \replabel{Reinflect-Verb} and (when using gold-standard labels for training, \Fref{fig:prefix_uriel}) the \replabel{WordLM} representations. However, with \replabel{WordLM} representations, the object and verb order as well as adposition and noun order features both explain the classification results about equally well (F$_1$ within 1.5 percentage points). For the \replabel{Reinflect-Verb} representations, the affixation type classification results can be explained by the negative affix position feature, which is not surprising given that it is included (along several other features) in the overall affixation position feature. The reinflection models have access only to word forms, without semantic or syntactic information, and so we do not expect them to differentiate between grammatical categories.
In addition to overall prefixing/suffixing tendency, the following features related to morphology were examined:
\begin{itemize}[noitemsep,topsep=10pt]
    \item Whether case affixes are prefixes or suffixes
    \item Whether negative affixes are prefixes or suffixes
    \item Whether plural affixes are prefixes or suffixes
    \item Whether possessive affixes are prefixes or suffixes
    \item Whether TAM affixes are prefixes or suffixes
    \item Existence of a negative affix
\end{itemize}
Some of them can be classified well using reinflection model representations, but they are all strongly correlated with each other and with the overall prefix/suffix feature, which is a weighted mean including most of the above features. This makes it difficult to conclusively determine which feature(s) a certain classifier has learned.

The \replabel{Encoder} model does have access to both word form and semantics, in the form of projected word embeddings. In \Fref{fig:neg_prefix_uriel} (whether negation is expressed with a prefix or a suffix) and \Fref{fig:poss_prefix_uriel} (whether a possessive prefix or suffix is used), we see that this model does not seem to encode the position of these specific features any more clearly than the reinflection models, which likely only achieve high classification accuracy due to correlation with the position of other affixes in the same language. One reason for this failure to encode morphological information is that the model is faced with the difficult task of encoding the representations of 18 million vocabulary items. Unlike the reinflection models, the encoder model does not have the opportunity to copy information, but must store a mapping within its rather limited number of parameters (565 thousand). In future work, it may be worth investigating a model that predicts the word embeddings, rather than the form, given the embedding and form of another member of the same paradigm. Such a model could extract encoded lexical information directly from the source embedding, and could focus on identifying morphological information.

In summary, our reinflection models seem to encode the overall tendency towards prefixing or suffixing, while no models are able to single out the position of affixes for specific grammatical categories.

\subsection{Naive cross-validation results}
\label{sec:naive}

To illustrate the effect of not following our cross-validation setup (\Fref{sec:crossvalidation}), we now compare \Fref{fig:naive_ov_uriel} (naive cross-validation) with \Fref{fig:ov_uriel} (linguistically sound cross-validation), and \Fref{fig:naive_adjn_uriel} (naive) with \Fref{fig:adjn_uriel} (sound). Clear detections, such as object/verb order with the \replabel{WordLM} representations, are not affected much by the cross-validation setup and result in accurate classifiers in both cases. Language representations with baseline-level results, such as our NMT-based models (\replabel{NMTeng2x} and \replabel{NMTx2eng}), perform equally poorly in both cases, suggesting that they do not correlate well with any type of language similarity. For representations such as \replabel{Lexical} and \replabel{ASJP}, the naive cross-validation setup results in much higher classification F$_1$ than the linguistically sound cross-validation. This is expected, since previous research has shown that the similarity metrics used to create these language representations can be used to reconstruct genealogical trees \citep{asjp}, which correlate well with typological features. The character-based language models (\replabel{Ö\&T} and \replabel{CharLM}) also show a similar increase in classification accuracy when naive cross-validation is used, which may indicate that they too use their language embeddings mainly to encode lexical similarity.

\subsection{Analysis of disagreements}
\label{sec:error-analysis}

For most classification experiments, we use URIEL data as a gold standard for both training and evaluation. However, for a few features we have access to projected labels. Here we apply these both as labels for training our classifiers, and as an additional source of information when analyzing the predictions of the classifiers we train.

To begin with, we compare the results when using URIEL labels for training (\Fref{fig:ov_uriel}) with using projected labels (\Fref{fig:ov_proj}). The overall results are very similar, which indicates that the projected labels are useful for learning this feature, even though they diverge somewhat from the gold standard URIEL labels.

For a more detailed view of the results, we show 3-way confusion matrices for a number of features in \Fref{tab:confusion}, summarizing the three sets of labels we have:
\begin{enumerate}
    \item Gold-standard URIEL labels (upper/lower matrix), index $i$
    \item Projected label (row), index $j$
    \item Predicted label from classifier (column), index $k$
\end{enumerate}

\begin{table}[]
    \centering
    \small
    \begin{tabular}{cccc}
        \toprule
        OV/VO & OV/VO & AdpN/NAdp & AdpN/NAdp  \\
        URIEL & projected & URIEL & projected \\[8pt]
$\left(
\begin{blockarray}{ cc }
\begin{block}{( cc )}
  54.6 & 0.4 \\
  9.9 & 0.2 \\
\end{block} \\
\begin{block}{( cc )}
  1.3 & 0 \\
  7.3 & 26.4 \\
\end{block}
\end{blockarray}
\right)$ & 
$\left(
\begin{blockarray}{ cc }
\begin{block}{( cc )}
  53.5 & 1.5 \\
  8.7 & 1.4 \\
\end{block} \\
\begin{block}{( cc )}
  1.3 & 0 \\
  3.9 & 29.8 \\
\end{block}
\end{blockarray}
\right)$ &
$\left(
\begin{blockarray}{ cc }
\begin{block}{( cc )}
35.8 & 5.4 \\
0.1 & 0.0 \\
\end{block} \\
\begin{block}{( cc )}
0.0 & 1.8 \\
5.5 & 51.3 \\
\end{block}
\end{blockarray}
\right)$ &
$\left(
\begin{blockarray}{ cc }
\begin{block}{( cc )}
37.5 & 4 \\
0.1 & 0.0 \\
\end{block} \\
\begin{block}{( cc )}
0.0 & 1.8 \\
5.8 & 51.1 \\
\end{block}
\end{blockarray}
\right)$ \\
        \midrule
        RelN/NRel & NumN/NNum & AdjN/NAdj & SV/VS \\
        URIEL & URIEL & URIEL & URIEL \\[8pt]
$\left(
\begin{blockarray}{ cc }
\begin{block}{( cc )}
15.2 & 0.1 \\
9.5 & 0.0 \\
\end{block} \\
\begin{block}{( cc )}
0.0 & 0.0 \\
29.6 & 45.5 \\
\end{block}
\end{blockarray}
\right)$ &
$\left(
\begin{blockarray}{ cc }
\begin{block}{( cc )}
44.4 & 10.8 \\
0.7 & 2.2 \\
\end{block} \\
\begin{block}{( cc )}
1.6 & 3.4 \\
8.5 & 28.3 \\
\end{block}
\end{blockarray}
\right)$ &
$\left(
\begin{blockarray}{ cc }
\begin{block}{( cc )}
29.0 & 4.9 \\
1.9 & 1.4 \\
\end{block} \\
\begin{block}{( cc )}
8.7 & 2.5 \\
20.8 & 30.7 \\
\end{block}
\end{blockarray}
\right)$ &
$\left(
\begin{blockarray}{ cc }
\begin{block}{( cc )}
75.1 & 14.7 \\
0.0 & 1.1 \\
\end{block} \\
\begin{block}{( cc )}
0.4 & 6.1 \\
0.2 & 2.2 \\
\end{block}
\end{blockarray}
\right)$ \\
    \bottomrule
    \end{tabular}
    \caption{3-way confusion matrices. We denote these matrices as $M_{i,j,k}$, where the sub-matrix $i$ indicates the URIEL label, row $j$ the projected label, and column $k$ the classifier output. These all refer to the \emph{evaluation} label. The header indicates whether URIEL or projected labels were used for \emph{training}.
    All numbers are percentages of language families with a certain combination of labels. Language families with more than one doculect in the data contribute to multiple counts, but each family has equal total weight.}
    \label{tab:confusion}
\end{table}

To begin with, we can compare the matrices obtained for \replabel{WordLM} when training on URIEL labels ($M^{\mathrm{URIEL (OV/VO)}}$, top left in \Fref{tab:confusion}) and with projected labels ($M^{\mathrm{projected (OV/VO)}}$, second from left).
If disagreements between the language representation-based classifiers and the typological databases were mainly due to differences between the Bible doculects and those used by the WALS and Ethonogue database compilers, we would have expected a much higher agreement between projected and classified labels. On the contrary, the mean F$_1$ is actually somewhat lower when evaluated against projected labels, even when projected labels are used for training (mean F$_1$ is 0.851, compared to 0.910 when evaluated against URIEL).

The same pattern is present for another feature, order of adposition and noun (\Fref{fig:adp}), with confusion matrices in \Fref{tab:confusion}.
The mean F$_1$ with respect to projected labels is nearly identical with URIEL-trained classifiers (0.887) as with
classifiers trained on projected labels (0.869). We see occasional examples of the opposite case, where the mean F$_1$ is somewhat higher when evaluated against the projected labels, but our conclusion is that actual linguistic differences between the Bible corpus and URIEL do not alone explain the cases where our classifiers differ from the URIEL classifications.

A somewhat different result is shown in \Fref{fig:numn} and \Fref{tab:confusion} for the order of numeral and noun. Here, the mean F$_1$ is considerably higher (0.763) when trained on projected labels than on URIEL labels (0.684), where both figures are evaluated with respect to URIEL labels. This could be partly due to the fact that the projected labels are available for more languages, and the mean number of language families for each training fold is higher (101.1) for the projected labels than for URIEL labels (60.9). Recall that only one randomly sampled doculect per family is represented in each training fold, so the number of families corresponds to the number of training fold data points. The mean F$_1$ is not substantially different (difference is less than one percentage point) when evaluated on projected instead of URIEL labels, and this applies for both sets of training labels, which speaks against the hypothesis that the URIEL and projected labels represent substantially different interpretations of the feature.

One notable property of the confusion matrices in \Fref{tab:confusion} is that $M_{0,1,1}$ and $M_{1,0,0}$ are generally very low, which means that when the projected feature value agrees with the classifier prediction, this consensus is very often correct according to URIEL. To quantify this, we can compute the F$_1$ for the subset of data where projected features and classifier predictions agree. \Fref{tab:agreeing} shows how the F$_1$ of \replabel{WordLM} increases drastically when we evaluate on this subset alone, sometimes reaching perfect or near-perfect scores.
\begin{table}[h]
\centering
\begin{tabular}{lrr}
    \toprule
    & \multicolumn{2}{c}{\textbf{Mean F$_1$ score}} \\
    \textbf{Feature} & \textbf{All doculects} & \textbf{Projected = Predicted} \\
    \midrule
    Order of adjective and noun & 0.639 & 0.880 \\
    Order of numeral and noun & 0.762 &  0.947 \\
    Order of relative clause and noun & 0.648 & 0.999 \\
    Order of adposition and noun & 0.866 & 1.000 \\
    Order of object and verb & 0.896 & 0.980 \\
    Order of subject and verb & 0.702 & 0.865 \\
    \bottomrule
\end{tabular}
\caption{Family-weighted mean F$_1$ scores of classifiers trained using \replabel{WordLM} representations. The columns give values using \textbf{All doculects}, or only those doculects where the projected and the classifier-predicted value agrees (\textbf{Projected = Predicted}).}
\label{tab:agreeing}
\end{table}

The only apparent disagreement for the order of adposition and noun turns out to be an error in URIEL.\footnote{Strangely, URIEL codes Serbian as having postpositions, even though \citet{wals-85} correctly codes it as prepositional.} For the order of object and verb, URIEL disagrees in five cases: Mbyá Guaraní (Tupian), Purépecha (isolate), Koreguaje (Tucanoan), Luwo (Nilotic), Yine (Arawakan). We have located grammatical descriptions in languages readable to us for three of these, in addition to quantitative word order data for Mbyá Guaraní.

\citet{choi-etal-2021-investigating} compare basic word order obtained from Universal Dependencies corpora \citep{ud22} with those in WALS \citep{wals} and \citet{Ostling2015wordorder}, and question the classification of Mbyá Guaraní as SVO-dominant since SOV is nearly as common.\footnote{\citet{choi-etal-2021-investigating} in fact compared Mbyá Guaraní with Paraguayan Guaraní (personal communication), which is coded as SVO by \citet{wals-81}, citing \citet[p 182]{GregoresSuarez1967} who describe Paraguayan Guaraní as having a rather free word order with SVO order being the most common, although they note that statements on word order should be taken as ``very rough approximations, based on impressionistic evaluations of what is more frequent.''}

Yine is classified by Ethnologue as an SOV language while our classification and projection methods both show a tendency towards VO order. \citet[p 292]{Hanson2010yine} states that ``The relative order of predicate and arguments varies considerably under pragmatic and stylistic motivations [...] The predicate-first order is somewhat more common than argument-first in verbal clauses.''

For Purépecha, \citet{wals-83} has SVO order. \citet[pp 61--62]{Friedrich-1984} gives SOV order but adds that ``the object-verb rule is weak.'' and further specifies that ``Short objects and, often, pronominal ones are generally preverbal. [...] Objects  with  two or more words, especially long words, tend  to be placed after the verb.'' There is no attempt at quantifying these statements.

From these examples, we see that when classifications from WALS or Ethnologue disagrees with a classifier/projection consensus with regards to verb/object order, in all cases we have investigated this can be attributed to the languages having a flexible word order, where the identification of a single dominant word order can be called into question.

Our interpretation of the generally high agreement when the classifier and projections agree is that these two methods, at least for our \replabel{WordLM} embeddings, complement each other. When both of them agree it is likely that the language is a clear example of the feature in question, and thus also likely to be classified as such by the database compilers. It is notable that we do \emph{not} see a corresponding improvement of classification performance in the subset of languages where URIEL and the projections agree, which again indicates the observed divergences can not only be explained by widespread grammatical differences between Bible doculects and URIEL sources.

In a few cases we observe the effects of different definitions of particular word order properties. The main exception to the pattern of high agreement between projected/classified consensus and URIEL classifications can be found for adjective/noun order, where 8.7\% of families are classified as adjective--noun by both the projection approach and the classifier, but are noun--adjective according to \citet{wals-87}. In this group we find several Romance languages. As discussed earlier, these tend to use adjective--noun order for a set of very common core adjectives, whereas noun--adjective is more productive but may be less common on a token level. For several other language families we also find examples where the order between \emph{core} adjective concepts and nouns differs from the order between Universal Dependencies \textsc{adj}-tagged words and nouns. However, a more careful analysis would be required to determine the cause of this discrepancy.

For the order of relative clause and noun, we see that the classifier has mediocre performance for the full sample but is near-perfect in the subset where projected and predicted labels agree. Looking at the full confusion matrix in \Fref{tab:confusion}, we see that the classifier is very good at classifying relative--noun languages, while the projection method instead excels at classifying noun--relative languages. This is mainly driven by the 29.6\% of language families that are classified as noun--relative order by both URIEL and the projection method, while the classifier gives relative--noun order. The features that best explain (in terms of highest mean F$_1$) the classifications of the relative/noun classifier, are adposition/noun, possessor/noun and object/verb order. This is not surprising, since relative--noun languages are overwhelmingly postpositional, object--verb and possessor--noun. If the classifier has learned to use one or more of these features as a proxy for relative/noun order, we would expect the languages misclassified as relative--noun to also be mainly postpositional, object--verb and possessor--noun. This is precisely what we find, whereas languages correctly classified as noun--relative are overwhelmingly prepositional, verb--object and noun--possessor. In combination with high accuracy of the projection method for noun--relative order, this causes the classifier/projection consensus to be in nearly perfect agreement with \citet{wals-87} but partly due to reasons not directly related to relative clauses.

\section{Conclusions}

Perhaps the most important result of our work is that typological generalizations \emph{can} be discovered by neural models solving NLP tasks, but only under certain circumstances. For word order features, the language representations from our multilingual word-based language model (\replabel{WordLM}) result in highly accurate classifiers for a range of word order features, close to the accuracy of various hand-crafted approaches in previous work \citep[Figure 9]{Ponti2019cl} as well as our projection-based approach (\Fref{sec:wordorderstatistics}). The general tendency of languages to be prefixing or suffixing does also appear to be discovered by our reinflection models.

Apart from these examples, we do not find any clear evidence of typological features encoded in the 12 sets of language representations we investigated. In most cases classification results were consistent with random labels. In some cases, such as the \replabel{WordLM} model being able to distinguish prefixing languages from suffixing, we show that the results can be better explained by the classifier learning a \emph{different} but correlated typological parameter.

Through the representations from the word-level language model and reinflection models, as well as our features obtained through annotation projection, we establish estimates for how well a number of typological features can be extracted from our data. No other language representations, including those from previous work, even come close to this level. From this we conclude that the models have not encoded any of the syntactic or morphological features in our study, nor language features sufficiently correlated with the features studied to create an accurate classifier for any of them. It would be theoretically possible that some of the features are encoded in some language representations, but in a way not classifiable using a logistic regression classifier. This would however be difficult to verify, and our results show that at least the word-level language model and reinflection models do encode features that are identifiable by a linear classifier.

Several previous authors have showed that vector similarity between some set of language representations has a similar structure to traditional phylogenetic trees constructed by historical linguists \citep{Ostling2017multilm,Oncevay2019towards,Tan2019multilingual}, or more generally cluster along  family lines \citep{tiedemann2018emerging,He2019syntactic}. While these observations are correct, they do not reveal much about whether linguistic \emph{generalizations} are made by the model and encoded in the language representations. 

Classification-based evaluations can be used to probe directly whether certain features are encoded in a set of language representations, assuming that correlations with genealogically and geographically close languages are properly controlled for.
In \Fref{sec:naive}, we showed that if care is not taken to make the testing set of each classifier model as independent as possible of the training set, it is very easy to obtain spurious results.
\citet{Malaviya2017learning} reported identifying features of syntax and phonology in the language representations from a multilingual NMT system, and \citet{Bjerva2018fromphonology} found features of syntax, morphology and phonology in the language representations from the mulitilingual language model of \citet{Ostling2017multilm}. Both relied on typological feature classification experiments. When strict separation of related languages between training and testing folds in the cross-validation is enforced, only a few solid identifications of typological features stand out, and these all come from our new models.
Both \citet{Malaviya2017learning} and \citet{Bjerva2018fromphonology} did take some precautions to avoid correlations between features of close languages affecting their results. However, even though the precise cause for the discrepancy between our respective conclusions have not been conclusively determined, we believe that our identification of typological generalizations by neural models is much more robust and unambiguous than in previous work. In some cases, the accuracy obtained by our classifiers even exceeds that of hand-coded annotation projection. This makes us able to not only demonstrate that neural models can discover typological features, but also that they can be used in practice to classify languages according to those features. When combining the results of the language representation-trained classifier and our projection method, the agreement with manually coded features can be even further increased. In part we believe this is due to the methods being complementary. Our word-based language model uses projected word embeddings and cosine loss in order to train efficiently with the full 18 million word vocabulary of all 1295 languages, and is not limited by the Universal Dependencies annotations that our projection method relies on.

One limitation of our study is that we only try to connect neural language representations with known features from the typological literature. But do the models also make typological generalizations that human linguists have \emph{not} previously made? We believe this would be an interesting topic for future research.

\section{Acknowledgments}

Thanks to Bernhard W{\"a}lchli, Mats Wir{\'e}n and Dmitry Nikolaev for valuable comments on this manuscript at  different stages. The computations were enabled by resources provided by the Swedish National Infrastructure for Computing (SNIC) at C3SE partially funded by the Swedish Research Council through grant agreement no.\ 2018-05973. This work was funded in part by the Swedish Research Council through grant agreement no.\ 2019-04129. 

\bibliographystyle{compling}
\bibliography{all}

\end{document}